\documentclass[sigconf,nonacm]{acmart}

\AtBeginDocument{
  }

\setcopyright{none}

\usepackage{listings}
\usepackage[most]{tcolorbox}
\usepackage{multirow}
\usepackage{CJKutf8}
\usepackage{subfigure}
\usepackage{caption}
\usepackage{float}
\usepackage{graphicx}

\begin{document}

\title{CFBenchmark-MM: Chinese Financial Assistant Benchmark for Multimodal Large Language Model}

\author{
  Jiangtong Li$^{*,1}$,
  Yiyun Zhu$^{*,1}$,
  Dawei Cheng$^1$, 
  Zhijun Ding$^1$,
  Changjun Jiang$^1$
}

\affiliation{
  \institution{
    $^*$ Both authors contributed equally to this research.}
  \country{
    $^1$ School of Computer Science and Technology, Tongji University}
}

\renewcommand{\shortauthors}{Li et al.}

\begin{abstract}
Multimodal Large Language Models (MLLMs) have rapidly evolved with the growth of Large Language Models~(LLMs) and are now applied in various fields. 
In finance, the integration of diverse modalities such as text, charts, and tables is crucial for accurate and efficient decision-making. 
Therefore, an effective evaluation system that incorporates these data types is essential for advancing financial application.
In this paper, we introduce CFBenchmark-MM, a Chinese multimodal financial benchmark with over 9,000 image-question pairs featuring tables, histogram charts, line charts, pie charts, and structural diagrams.
Additionally, we develop a staged evaluation system to assess MLLMs in handling multimodal information by providing different visual content step by step.
Despite MLLMs having inherent financial knowledge, experimental results still show limited efficiency and robustness in handling multimodal financial context. 
Further analysis on incorrect responses reveals the misinterpretation of visual content and the misunderstanding of financial concepts are the primary issues. 
Our research validates the significant, yet underexploited, potential of MLLMs in financial analysis, highlighting the need for further development and domain-specific optimization to encourage the enhanced use in financial domain.
\end{abstract}



\maketitle

\section{Introduction}
In recent years, MLLMs such as GPT-4V~\cite{GPT-4}, Gemini~\cite{Gemini}, and Claude~\cite{Claude3} have grown significantly, excelling at integrating text and visuals for processing and generating information. Similarly, open-source MLLMs like LLaVA~\cite{liu2024visual}, CogVLM~\cite{wang2023cogvlm}, and InternLM-XComposer~\cite{dong2024internlm2} advance by leveraging the foundations of LLMs to enhance visual content integration across various domains.

In financial domain, analyzing charts such as stock trends, financial statements, and economic data visualizations becomes crucial for decision-making, which relies on both structured and unstructured data. 
Therefore, MLLMs need to manage complex financial context that combine textual data with visual charts, requiring high levels of analysis and comprehension. 
However, a comprehensive benchmark to evaluate MLLMs in handling complex multimodal financial context is still absence.

Early financial benchmarks~\cite{shah2022flue, chen2023disc} focus on specific tasks like sentiment analysis of news~\cite{mishev2020evaluation,sinha2022sentfin} or statistics reasoning~\cite{chen2021finqa,shah2022flue}, evaluating single model performance within narrow domains. 
As benchmarks evolve~\cite{lu2023bbt, zhang2023fineval, xie2024finben}, they begin assessing the generalization of models across diverse professional tasks. 
Recent developments in benchmarks~\cite{bhatia2024fintral} include non-text data like charts, enhancing realism and accuracy in multimodal scenarios. However, most financial benchmarks~\cite{shah2022flue,bhatia2024fintral} still either concentrate solely on text or are limited to specific multimodal tasks such as stock tracking. 
This underscores the need for a more comprehensive multimodal financial benchmark that not only evaluates financial knowledge and analytical skills but also measures effectiveness in utilizing and integrating multimodal information from various visual formats like pie charts, line charts, tables, and structural diagrams.
\begin{figure*}[t]
    \centering
    \includegraphics[width=1\textwidth]{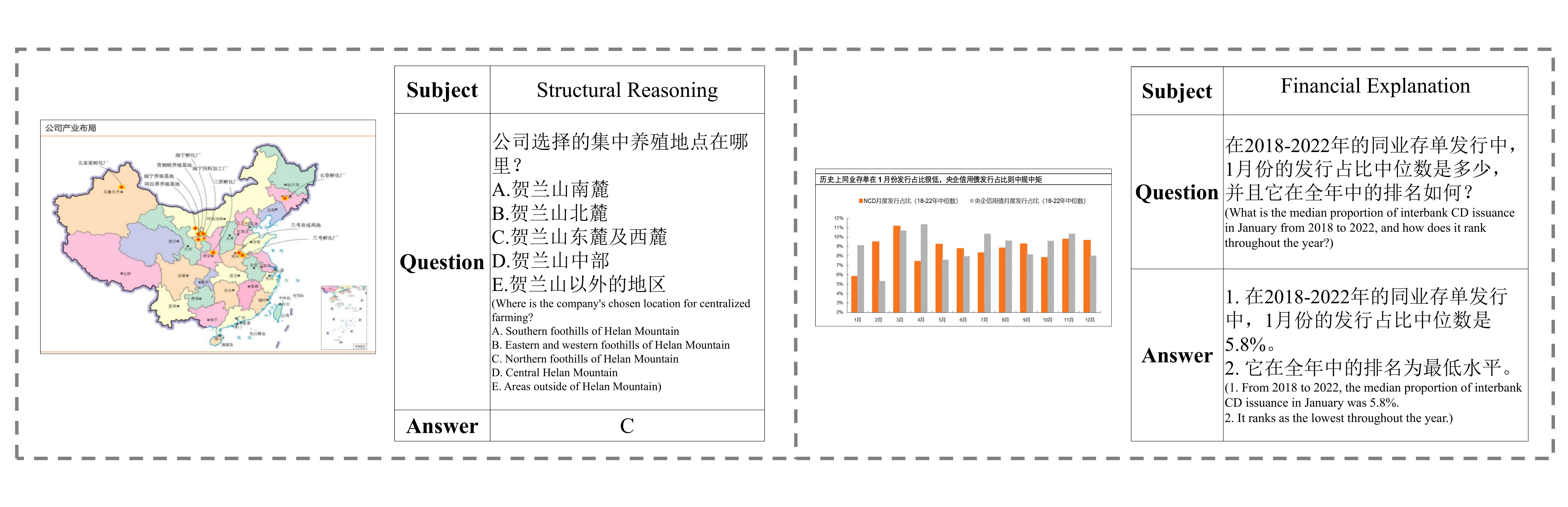}
    \caption{Two examples from CFBenchmark-MM.}
    \label{fig:cases}
\end{figure*}
In this paper, we introduce CFBenchmark-MM, which bridges gaps in existing benchmarks, providing a detailed and realistic evaluation for MLLMs in complex financial context with diverse data types. 
Specifically, CFBenchmark-MM includes five core tasks, \emph{i.e.}, arithmetic reasoning, statistical reasoning, structural reasoning, financial explanation, and financial knowledg, with 9,356 image-question pairs. 
To construct CFBenchmark-MM, we develop a three-step annotation pipeline: 1) we collect charts and their analyses from financial research reports; 2) we exploit GPT-4 to generate questions and answers based on the text analyses; 3) we manually verify whether the questions can be answered using the charts alone. 
In Figure~\ref{fig:cases}, we present two cases from CFBenchmark-MM.

To assess the capability and efficiency of MLLMs in handling multimodal financial information, we develop a staged evaluation system including multiple settings. 
Initially, MLLMs answer questions without visual content to evaluate their basic memory. Progressively, they respond based on a) textual descriptions of charts, b) charts, and c) both charts and their textual descriptions. 
This staged approach helps measure how MLLMs is enhanced with multimodal financial information in answering questions. 
As for the evaluation metrics, we employ a multiple-choice format, measuring accuracy, for objective questions in arithmetic, statistical, and structural reasoning. 
And we compare the generated answers to reference answers, using GPT-4 as a scorer, for subjective financial explanation and knowledge questions. 
Scores reflect the proportion of key information matched.

Our experiments reveal that most MLLMs have a fundamental memory of financial knowledge, allowing them to answer questions such as "What is the increase rate of Chinese GDP over 2022?" or "Which stock grew the second largest in 2021?" 
Including multimodal information improves response accuracy, yet overall performance remains moderate. 
Even the advanced GPT-4V only achieves a 52\% accuracy and 38\% scores on objective and subjective questions, highlighting the need for further enhancements in financial MLLMs. 
Additionally, our analysis of incorrect responses pinpoints that misinterpretation of visual content and misunderstanding of financial concepts are the primary issues. 
These findings emphasize the importance of MLLMs to analyze multimodal content in financial contexts.
Our contributions can be summarized as: 

\begin{itemize}
    \item We introduce CFBenchmark-MM, a comprehensive benchmark with over 9,000 image-question pairs, focusing on evaluate MLLMs in real multimodal financial context.
    \item We establish an elaborated annotation process using real multimodal financial data, which includes automated tagging and manual verification to ensure dataset quality.
    \item We develop a staged evaluation system with multiple settings to assess the financial abilities of MLLMs and their use of multimodal information step by step, providing insights into existing challenges and guiding future enhancements.
\end{itemize}

\section{Related Work}

\subsection{Large Language Models in Finance}

Large Language Models (LLMs)~\cite{raffel2020exploring, touvron2023llama, du2021glm, dettmers2024qlora, yang2019xlnet, chang2024survey}, such as the GPT series by OpenAI~\cite{radford2018improving, radford2019language, brown2020language,GPT-4}, feature billions of parameters and extensive training on text data.
Notable among them, BloombergGPT~\cite{wu2023bloomberggpt} stands out as a FinLLM with 50 billion parameters. 
Successor models like FinGPT~\cite{yang2023fingpt} and PIXIU~\cite{xie2023pixiu} improve performance through supervised fine-tuning, excelling in zero-shot or few-shot settings. 
In the Chinese financial domain, XuanYuan~\cite{zhang2023xuanyuan} specializes in question-answering with domain-specific fine-tuning, while DISC-FinLLM~\cite{chen2023disc} enhances reasoning in finance using multi-expert fine-tuning with LoRA~\cite{hu2021lora}. Despite these advancements, current FinLLMs show a limited focus on multimodal data. 
Thus, this paper focuses on multimodal evaluation in finance, aiming to guide the development of MLLMs in this domain.

\subsection{Multimodal Large Language Models}
MLLMs~\cite{yin2023survey}, evolving from LLMs, focus on integrating information from various modalities with large language models, which can be divided into pre-training for visual-language alignment and fine-tuning for visual instructions. 
For visual-language pre-training methods, models like CLIP~\cite{radford2021learning} use dual encoders to evaluate matches between images and text, while others like ALBEF~\cite{li2021align} and VLMo~\cite{bao2022vlmo} enhance interaction between these modalities. 
BLIP~\cite{li2022blip} and its successor, BLIP-2~\cite{li2023blip}, improve modality integration with query transformers and noise reduction, and CogVLM~\cite{wang2023cogvlm} introduces visual expert modules for deeper alignment.
For visual-instruction fine-tuning methods, models such as InstructBLIP~\cite{dai2024instructblip} and LLaVA~\cite{liu2024visual} enhance instruction awareness and align image features with linguistic embeddings, respectively. LaVIT~\cite{jin2023unified} and MiniGPT-4~\cite{zhu2023minigpt} process visual and language data by converting visuals into discrete tokens and adjusting pre-trained models for multimodal instructions. 
The development of models like GPT-4~\cite{GPT-4} showcases advanced capabilities in handling visual tasks and improving multimodal interaction across various benchmarks.

Despite the progress made by MLLMs, research on multimodal charts in financial domain remains sparse. 
Therefore, this work focuses on multimodal evaluation in real financial contexts, aiming to enhance the development of MLLMs specifically tailored for the financial domain.

\subsection{Benchmarks}

Research on evaluation benchmarks is essential for developing specific domains and provides a crucial framework for assessing the performance of large models.
In this section, we discuss the evolution of benchmarks in financial domains for MLLMs, highlighting their significance and impact.
In financial domain, benchmarks like FLUE~\cite{shah2022flue}, CFLEB~\cite{lu2023bbt}, and DISC-FinLLM~\cite{chen2023disc} evaluate NLP capabilities, while FinEval~\cite{zhang2023fineval} ,investorbench~\cite{li2024investorbenchbenchmarkfinancialdecisionmaking} and FinQA~\cite{chen2021finqa} focus on financial analysis and knowledge. 
Other benchmarks such as FinBen~\cite{xie2024finben}, SuperCLUE-Fin~\cite{xu2024superclue}, and FinanceIQ~\cite{zhang2023xuanyuan} merge NLP and financial skills assessment across various applications, significantly advancing LLMs in the financial domain. 
However, their limited focus on multimodal domains does not fully meet the practical requirements of financial applications.
In general domains, several authoritative benchmarks~\cite{ye2023mplug, yu2023mm, liang2021multibench, li2021value,yue2024mmmumassivemultidisciplinemultimodal,liu2024mmbenchmultimodalmodelallaround} exist for multimodal evaluation. 
MME~\cite{fu2024mme} is a widely used benchmark for evaluating MLLMs, assessing 14 sub-tasks of perceptual and cognitive abilities. 
ShareGPT4V~\cite{chen2023sharegpt4v} includes 1.2M image-text pairs, from diverse aspects such as world knowledge and artistic evaluations. Other multimodal evaluation like MME-industry~\cite{yi2025mmeindustrycrossindustrymultimodalevaluation} focus on specific domain.
These benchmarks provide comprehensive evaluations and highlight improvement of MLLMs in general domain.

However, few studies, like FinVQA~\cite{bhatia2024fintral}and MME-Finance~\cite{gan2024mmefinancemultimodalfinancebenchmark} focus on multimodal evaluation in finance. 
Specifically, FinVQA assesses how models handle different financial modalities but is limited by its small dataset size and narrow scope. MME-Finance proposes a bilingual financial multimodal benchmark, but also has the limitation of insufficient amount. This underscores the need to develop comprehensive multimodal benchmarks to thoroughly evaluate MLLMs in managing real financial context.
\section{Methodology}

In order to construct a multimodal evaluation benchmark that aligns with the practical requirements of financial application, we engage with professionals from the different financial domain, including consultants, bankers, investment bankers, and securities experts. 
Through extensive discussions with these industry specialists, we design and analyze our CFBenchmark-MM from two key dimensions: the types of questions and the types of multimodal data.

In terms of the question type, we classify the questions in our benchmark into five categories, which includes three categories involving objective evaluations, \emph{i.e.}, \textbf{Arithmetic Reasoning}, \textbf{Statistical Reasoning}, and \textbf{Structural Reasoning}, and two categories involving subjective evaluations, \emph{i.e.}, \textbf{Financial Explanation} and \textbf{Financial Knowledge}. 
Regarding the multimodal data, we further category them into various types such as tables, histogram charts, line charts, pie charts, composite charts, and structural diagrams (such as equity structure charts and development path charts). 
This classification is carefully crafted to ensure that our benchmark comprehensively covers the diverse needs and challenges encountered in multimodal financial applications.

Here is a precise definition of the five question types:
\begin{enumerate}
   \item \textbf{Arithmetic Reasoning}: This category evaluates the MLLMs in basic arithmetic operations such as addition, subtraction, multiplication, and division, as well as the understanding of numerical properties and relationships within financial charts.

   \item \textbf{Statistical Reasoning}: This category involves to interpret and analyze statistical data in financial domain. It focuses on statistical measures such as mean, median, mode, variance, standard deviation, and correlation coefficients in financial charts.

   \item \textbf{Structural Reasoning}: This category pertains to interpret and analyze structural diagrams and flowcharts in financial domain. This includes assessing diagrams that illustrate complex relationships such as shareholding structures or company development pathways.

   \item \textbf{Financial Explanation}: This category entails to interpret and analyze the financial trends, fluctuations, and anomalies. It requires synthesizing multimodal information to provide comprehensive explanations of financial phenomena and their underlying causes.

   \item \textbf{Financial Knowledge}: This category concerns the application and extension of financial knowledge through the analysis of multimodal data. It challenges the MLLMs to integrate and extrapolate given information in understanding financial concepts.
\end{enumerate}

\subsection{Benchmark Construction}

\begin{figure*}[ht] 
    \centering
    \includegraphics[width=1\linewidth]{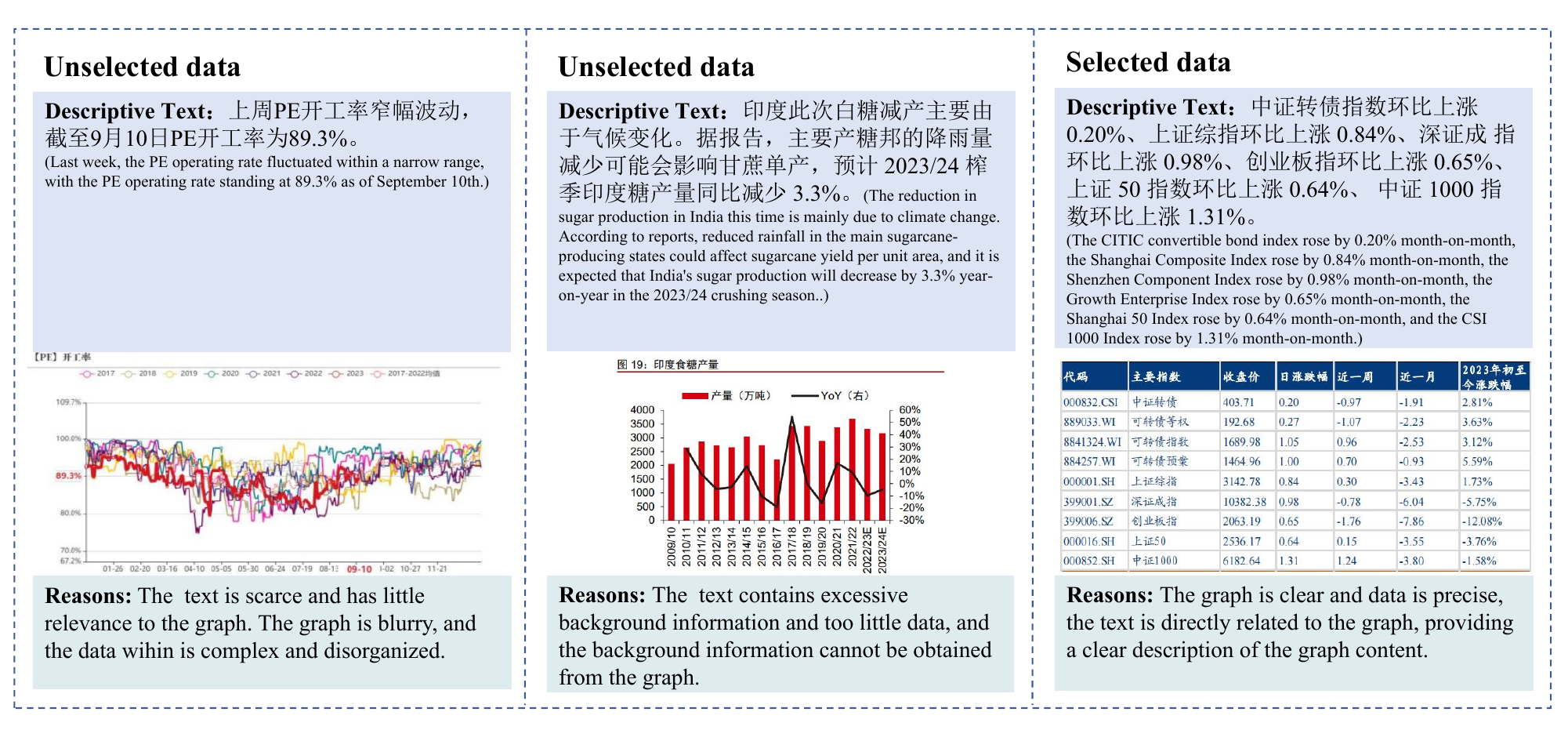}
    \caption{The criteria we use when selecting data from financial research reports. We choose financial data that are clear, precise, and contain analyzable content.}
    \label{fig:data_collection}
\end{figure*}

\subsubsection{Data Collection}

To construct a comprehensive benchmark, we first collect multimodal charts from publicly available financial research reports, which reflect actual usage patterns within the financial industry, providing an authentic basis for our evaluations.
Besides, during benchmark construction from on these charts, we have two options, 1) employ annotators to propose questions and answers directly from the chart content; 2) extract the analyses of charts from the reports and automatically generate corresponding questions and answers. 
However, the former method requires substantial manpower and could introduce annotator biases. 
To align with the actual analysis methods used in financial application, we opt to collect the textual analyses accompanying these multimodal charts. 
These textual analyses, while not encompassing all information presented in the charts, represent the most critical and significant insights according to the use cases of these reports.

To ensure the quality of the collected charts and their corresponding textual analyses, we establish specific screening criteria, 1) information in the charts is clear and visible without any obstructions, blurriness, or disorder; 2) the textual analysis hasto correspond closely with the information in the charts and be relatively extensive (exceeding 50 Chinese characters). In Figure~\ref{fig:data_collection}, we present several selected and unselected data to further explain our criteria.
Based on these principles, we organize each collection group with two undergraduate students. 
Only when both individuals agree on the clarity and relevance of a chart and its analysis, the collect charts-analysis pair is included. 
Through this meticulous approach, we compile a total of 2500 pairs of charts-analysis.

\subsubsection{Data Annotation} 

To ensure the quality of our benchmark, we utilize the advanced LLM, GPT-4, to generate questions and answers from the textual analysis of collected chart-analysis pairs. 
Based on the characteristics of the charts, the association between chart types and question categories are defined as 1) line charts and tables are used to generate \textbf{Arithmetic Reasoning} and \textbf{Financial Knowledge} question, 2) histogram charts and other types like pie charts and composite charts are used to generate \textbf{Statistical Reasoning} and \textbf{Financial Explanation} questions, 3) structural diagrams are used to generate \textbf{Structural Reasoning} and \textbf{Financial Explanation} questions.
While proposing the questions and answers, we design five specific instructions for GPT-4, which require the model to first assess if the provided analysis is suitable for question generation. 
If deemed appropriate, the model then proceed to craft questions aligned with the requirements of each of the five question categories. 
More details about these instructions are available in appendix. 
For each analysis that qualified, the model is instructed to generate two questions per question category, ensuring a comprehensive assessment.

For subjective questions~(financial knowledge and explanation), the answers are formatted in bullet points to facilitate clarity, while for objective questions~(arithmetic, statistical, and structural reasoning), the answers are structured as multiple-choice. 
Through this methodical approach, we successfully identified 2,421 chart-analysis pairs suitable for this exercise, ultimately generating a robust dataset of 9,684 image-question pairs. 
This annotation procedure ensures the quality of question-answer pairs and their relation with financial charts and provides solid support for evaluating MLLMs in multimodal financial context.

\subsubsection{Manual Examination} 

After completing data annotation, we conduct manual verification through the following steps, 1) we verify whether the proposed questions can be answered using only the provided multimodal information; 2) for objective questions, checking the correctness of the option format and ensuring that the correct answer is included among the options; 3) for subjective questions, confirming the accuracy of answers formatted in bullet points; 4) we manually label questions that are unrelated to the charts or where the correct information cannot be obtained from the charts.
After the manual examination, we finally obtain 2,339 valid charts along with 9,356 questions to construct CFBenchmark-MM.

\begin{figure*}[ht]
    \centering
    \includegraphics[width=1\linewidth]{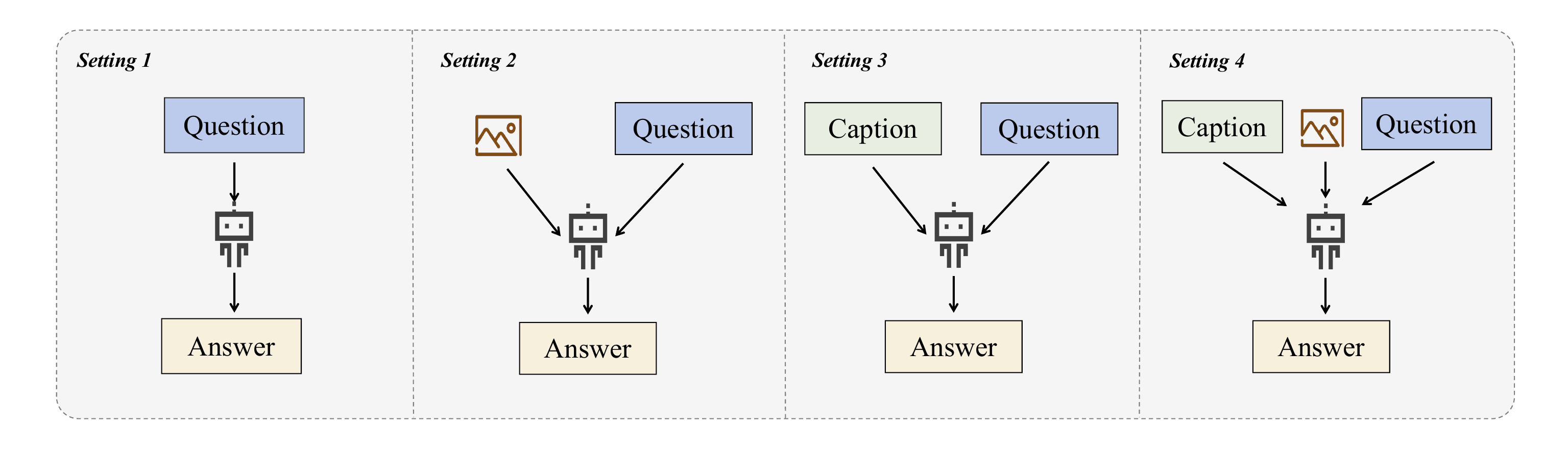}
    \caption{The illustration of benchmark evaluation system.}
    \label{fig:benchmark-evaluation}
\end{figure*}

\begin{figure}[t]
    \centering
    \includegraphics[width=0.5\linewidth]{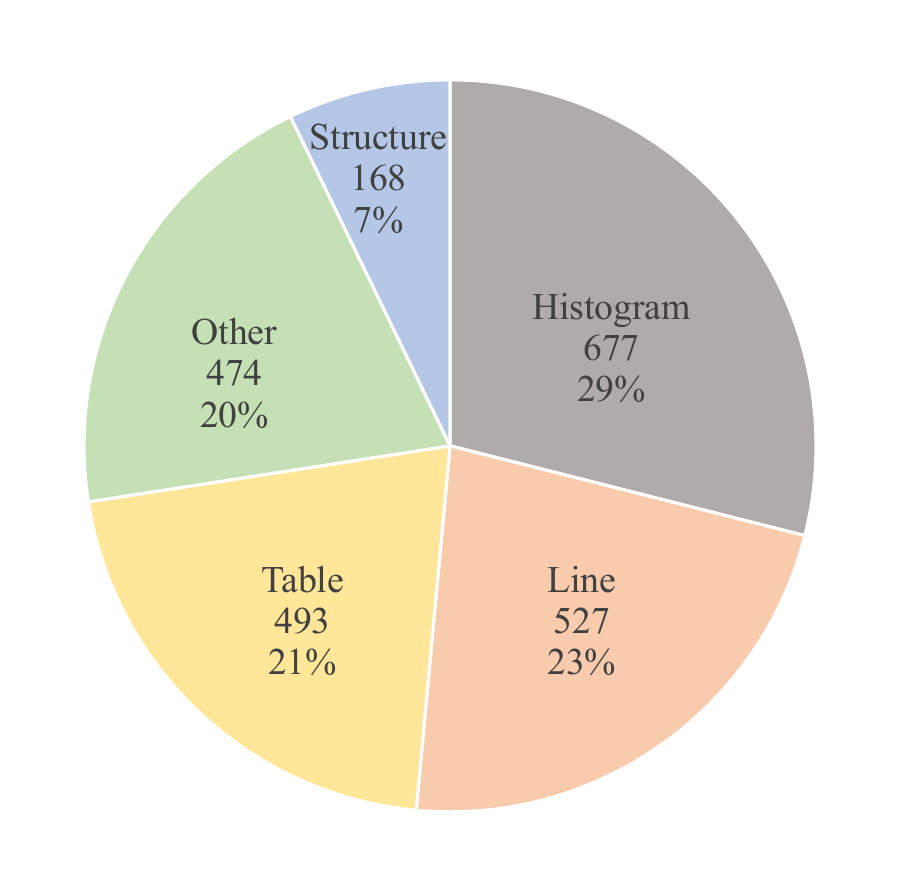}
    \caption{The distribution of different types of charts in CFBenchmark-MM.}
    \label{fig:pi_statistics}
\end{figure}

\subsubsection{Statistics}

In Figure~\ref{fig:pi_statistics}, we display the distribution of different types of charts used in CFBenchmark-MM, where the most common type, histogram charts, totals 677 charts, accounting for 29\%, and the least common, structural diagram, includes only 168 charts, making up 7\%. 
This variation reflects the diversity in financial research reports. 
In Table~\ref{tab:question_statistics}, we present the statistics for different categories of questions.
Financial explanation questions are most frequent due to the higher availability of corresponding charts, while structural reasoning questions are least frequent, reflecting the scarcity of structural diagrams. 
Additionally, subject questions often feature multiple correct answers, resulting in answers longer than one word. 
Our CFBenchmark-MM, with its extensive range and high volume, is the most comprehensive to date and plays a significant role in multimodal financial context.

\subsection{Benchmark Evaluation}

Most existing MLLMs have been trained on financial text data and may inherently possess some capabilities to answer benchmark-related questions. 
To fully evaluate the financial and multimodal processing capabilities, we develop a staged evaluation system with multiple settings for CFBenchmark-MM. 
In Figure~\ref{fig:benchmark-evaluation}, we illustrate the procedure of our staged evaluation system, as detailed below:

\begin{enumerate}
  \item \textbf{Question~(setting-Q):} This baseline evaluation assesses models MLLMs on questions without additional multimodal information. It reflects the inherent knowledge of MLLMs.
  
  \item \textbf{Question + Image~(setting-Q+I):} In this setting, we introduce original charts into the evaluation to examine how models integrate visual content with textual queries, testing their multimodal processing capabilities.
  
  \item \textbf{Question + Caption~(setting-Q+C):} In this setting, we enhance the assessment by adding textual descriptions generated by advanced MLLM, GPT-4V, from the charts, evaluating how well models can utilize synthesized textual representations alongside the queries.
  
  \item \textbf{Question + Image + Caption~(setting-Q+I+C):} This setting combines both the original visual content and its textual description into assessment, measuring the ability of MLLMs to synergize and scale their understanding across complex multimodal inputs.
\end{enumerate}

Through this staged evaluation, we not only gauge the foundational knowledge of MLLMs in financial domain but also critically assess their proficiency in handling both direct multimodal inputs and derived textual representations. 
This evaluation strategy provides valuable insights into model performance, aiding in the effective selection of models for real-world financial applications.

\begin{table}[t]
    \resizebox{\linewidth}{!}{
    \setlength\tabcolsep{4pt}
    \begin{tabular}{ccccc}
    \hline \hline
    Category                                                         & \#Chart & \#Question & Avg. Length (Q) & Avg. Length (A) \\ \hline \hline
    \begin{tabular}[c]{@{}c@{}}Arithmetic \\ Reasoning\end{tabular}  & 1020 & 2040 & 79.7 & 1.1 \\ \hline
    \begin{tabular}[c]{@{}c@{}}Statistical \\ Reasoning\end{tabular} & 1151 & 2302 & 69.9 & 1.4 \\ \hline
    \begin{tabular}[c]{@{}c@{}}Structural \\ Reasoning\end{tabular}  & 168  & 336  & 82   & 1.2 \\ \hline
    \begin{tabular}[c]{@{}c@{}}Financial \\ Explanation\end{tabular} & 1319 & 2638 & 33.9 & 110.8 \\ \hline
    \begin{tabular}[c]{@{}c@{}}Financial \\ Knowledge\end{tabular}   & 1020 & 2040 & 28.4 & 275.3 \\ \hline \hline
    \end{tabular}
    }
    \caption{The statistics about CFBenchmark-MM in terms of question category.}
    \label{tab:question_statistics}
\end{table}

\begin{figure*}[ht]
    \centering
    \includegraphics[width=1\linewidth]{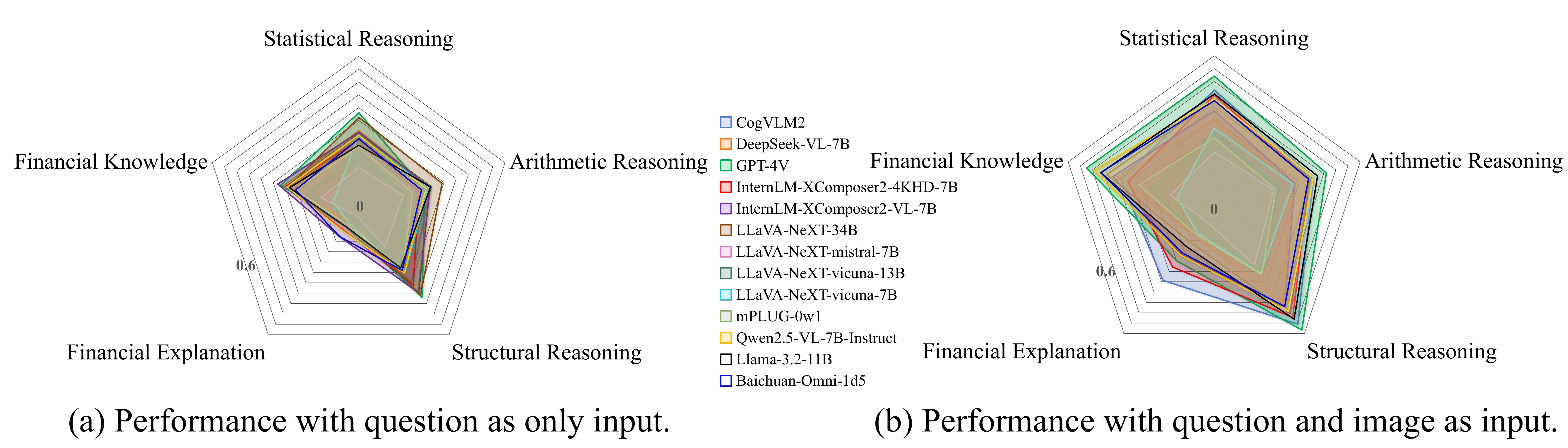}
    \caption{We evaluate 13 cutting-edge MLLMs on CFBenchmark-MM with question and with (question, image) as input}
    \label{fig:radio_graph}
\end{figure*}

For evaluation metric, objective questions are assessed through a multiple-choice format, with accuracy determined by correct options. 
Subjective questions, like financial explanation and knowledge, are evaluated by matching generated answers to reference answers using GPT-4 as a scorer. 
Scoring reflects the proportion of key points accurately captured. 
For instance, if three out of five key points in a reference answer are matched, the score is 0.6. `

\section{Experiments}

\subsection{Experimental Setup}

We evaluate the zero-shot performance of 13 cutting-edged MLLMs: (1) LLaVA-NeXT-mistral-7B~\cite{liu2024visual}; (2) LLaVA-NeXT-vicuna-7B~\cite{liu2024visual}; (3) LLaVA-NeXT-vicuna-13B~\cite{liu2024visual}; (4) LLaVA-NeXT-34B~\cite{liu2024visual}; (5) CogVLM2~\cite{wang2023cogvlm}; (6) DeepSeek-VL-7B~\cite{lu2024deepseek}; (7) InternLM-XComposer2-4KHD-7B~\cite{dong2024internlm2}; (8) InternLM-XComposer2-VL-7B~\cite{dong2024internlm1}; (9) mPLUG-Owl~\cite{ye2023mplug}; (10) GPT-4V~\cite{GPT-4}; (11) Llama-3.2-11B-Vision~\cite{grattafiori2024llama3herdmodels}; (12) Qwen2.5-VL-7B-Instruct~\cite{bai2025qwen25vltechnicalreport};(13) Baichuan-Omni-1d5~\cite{li2025baichuanomni15technicalreport}.
For open-source MLLMs, we directly utilize their released model weights on single Nvidia A100 GPU during the evaluation.
For GPT-4V, we utilize the API for evaluation.
Moreover, some popular MLLMs, like, BLIP~\cite{li2022blip, li2023blip}, and InstructBLIP~\cite{dai2024instructblip}, cannot be evaluated in our benchmark due to the language support.
The instruction used during the evaluation is shown below.

\begin{tcolorbox}[auto outer arc, boxrule=0.5pt, colback=white!90!yellow, left=1mm, right=1mm, top=1mm, bottom=1mm]
\scriptsize
\begin{CJK}{UTF8}{gbsn}
以下是一道关于\{subject\}的题目，一张金融场景中的图表和对应的文字描述，请基于这张图表和文字描述回答问题。图表和文字描述可能为空，如果其为空时，请忽略其内容。这道题可能是多项选择题，也可能是简答题。针对多项选择题，请你选出所有的正确选项；针对简单题，请你逐点回答这个问题。这张图表可能是表格、结构图、饼图、折线图或柱状图。请在回答时仔细浏览图表的每个组成部分，包括但不限于：标题、X轴和Y轴的标签、图例说明、各个数据系列的颜色和形状、图表中显示的数据点和数值。如果是表格，请详细浏览的标题和内容。如果是饼图，请说明每个扇区的百分比和扇区名称。如果数值没有标注在图上，请你根据X轴和Y轴判断数值大小，并基于判断的数值大小进行回答。题目为\{question\}。图表为\{<IMAGE>\}。图表描述为\{caption\}。
\end{CJK}\\
(Here is a question about \{subject\}, featuring a chart from financial domain and its corresponding textual description. Based on the chart and text provided, please answer the question. The chart and text may be absent; if so, ignore their content. The question may be multiple-choice or a short-answer type. For multiple-choice questions, select all correct options; for short-answer questions, respond to each point in the question. The chart could be a table, structural diagram, pie chart, line chart, or bar chart. When answering, carefully examine each component of the chart, including but not limited to: the title, labels on the X and Y axes, the legend, the colors and shapes of data series, and the data points and values displayed. If it is a table, closely review the title and content. If it is pie charts, specify the percentage and name of each sector. If values are not marked on chart, determine the size of the values based on the X, Y axes and answer based on these estimated values.
The question is \{question\}. The chart is \{<IMAGE>\}. The chart description is \{caption\}.)
\normalsize
\end{tcolorbox}

\subsection{Main Results}

In Figure~\ref{fig:radio_graph}, we display the performance of 13 MLLMs under setting-Q and setting-Q+I, showcasing their capabilities within multimodal financial contexts. Despite only having questions as input, most MLLMs achieve over 20\% accuracy ~(accuracy for random choice) in objective questions and score above 10\% in subjective assessments, demonstrating their foundational memory in financial knowledge. With the addition of original charts (setting-Q+I), all MLLMs exhibit notable improvements, indicating the value of integrating visual data for enhanced comprehension. Among them, GPT-4V stands out, achieving the largest improvement upon integrating charts, with over 15\% accuracy and 20\% score increases in objective and subjective questions, respectively. This underscores its superior multimodal processing capabilities. In contrast, LLaVA-NeXT-mistral-7B shows the weakest overall performance, with minimal improvement upon integrating charts, likely due to its limited Chinese language processing capabilities. However, even advanced MLLMs like GPT-4V achieve only 52\% accuracy and 38\% score in objective and subjective questions, falling short of real-world application requirements. Moreover, under both evaluation settings, most models perform best on Structural Reasoning questions, while their performance is weakest on tasks related to Financial Explanation. These findings highlight a substantial opportunity for further enhancements in multimodal financial applications, emphasizing the need for exploiting multimodal data in financial analysis and bridging the gap between capabilities and practical utility.

\begin{table*}[t]
\begin{center}
\resizebox{\textwidth}{!}{
\setlength\tabcolsep{8pt}
\begin{tabular}{c|cccc|cccc|cccc}
\hline\hline
\multirow{2}{*}{Question} & \multicolumn{4}{c|}{GPT-4V} & \multicolumn{4}{c}{LLaVA-NeXT-34B} & \multicolumn{4}{|c}{InternLM-X2-4K-7B} \\ \cline{2-13} 
 & Q & Q+C & Q+I & Q+C+I & Q & Q+C & Q+I & Q+C+I & Q & Q+C & Q+I & Q+C+I \\ \hline
\multirow{2}{*}{Arithmetic} & 26.9 & 35.4 & \textbf{46.0} & 38.1 & 34.0 & 38.0 & 39.2 & 42.1 & 24.4 & 32.8 & 38.8 & 40.1 \\
 & - & +8.5 & \textbf{+19.1} & +11.2 & - & +4.0 & +5.2 & +8.1 & - & +8.4 & +14.4 & +15.7 \\
\multirow{2}{*}{Statistical} & 38.0 & 41.5 & \textbf{52.1} & 45.2 & 36.2 & 40.2 & 42.8 & 45.2 & 27.4 & 36.6 & 44.6 & 45.9 \\
 & - & +3.5 & +14.1 & +7.2 & - & +4.0 & +6.6 & +9.0 & - & +9.2 & +14.2 & \textbf{+15.5} \\
\multirow{2}{*}{Structural} & 41.9 & 50.1 & \textbf{58.1} & 54.6 & 40.8 & 47.1 & 50.3 & 55.7 & 35.9 & 47.3 & 52.1 & 54.2 \\
 & - & +8.2 & +16.2 & +12.7 & - & +6.3 & +9.5 & +15.7 & - & +11.4 & +16.2 & \textbf{+18.1} \\
\multirow{2}{*}{Explanation} & 8.2 & 11.5 & 24.8 & 23.8 & 6.3 & 13.6 & 15.8 & 17.1 & 9.8 & 12.9 & 27.7 & \textbf{31.1} \\
 & - & +3.3 & +16.6 & +15.6 & - & +7.3 & +9.5 & +10.8 & - & +3.1 & +17.9 & \textbf{+21.3} \\
\multirow{2}{*}{Knowledge} & 31.8 & 38.1 & \textbf{52.4} & 49.9 & 29.0 & 34.6 & 44.2 & 48.3 & 29.9 & 33.5 & 35.5 & 38.9 \\
 & - & +6.3 & \textbf{+20.6} & +18.1 & - & +5.6 & +15.2 & +19.3 & - & +3.6 & +5.6 & +9.0 \\
\multirow{2}{*}{Average} & 29.4 & 35.3 & \textbf{46.7} & 42.3 & 29.3 & 34.7 & 38.5 & 41.7 & 25.5 & 32.6 & 39.7 & 42.0 \\
 & - & +5.9 & \textbf{+17.3} & +12.9 & - & +5.4 & +9.2 & +12.4 & - & +7.1 & +14.2 & +16.5 \\ \hline\hline
\end{tabular}
}
\end{center}
\caption{Evaluation results on GPT-4V, LLaVA-NeXT-34B, and InterLM-XComposer2-4KHD-7B~(InternLM-X2-4K-7B for short) in different evaluation settings. Q, C, and I indicate question, caption, and image, respectively. The best performance in different tasks are highlight in bold.}
\label{tab:3model_compare}
\end{table*}

\begin{table*}[ht]
\begin{center}
\resizebox{\textwidth}{!}{
\setlength\tabcolsep{11pt}
\begin{tabular}{c|ccc|ccc|ccc}
\hline \hline
\multirow{2}{*}{Question} & \multicolumn{3}{c|}{GPT-4V} & \multicolumn{3}{c|}{LLaVA-NeXT-34B} & \multicolumn{3}{c}{InternLM-X2-4K-7B} \\ \cline{2-10} 
 & Original & Simple & ICL & Original & Simple & ICL & Original & Simple & ICL \\ \hline
Arithmetic  & 46.0 & 46.2 & 49.1 & 39.2 & 38.7 & 41.9 & 38.8 & 38.2 & 40.3 \\
Statistical & 52.1 & 51.7 & 54.8 & 42.8 & 42.5 & 44.3 & 44.6 & 43.9 & 46.1 \\
Structural  & 58.1 & 57.6 & 60.9 & 50.3 & 49.2 & 53.1 & 52.1 & 51.8 & 53.2 \\
Explanation & 24.8 & 24.4 & 28.3 & 15.8 & 14.8 & 18.4 & 27.7 & 27.3 & 29.2 \\
Knowledge   & 52.4 & 51.1 & 57.2 & 44.2 & 44.6 & 50.2 & 35.5 & 34.9 & 42.1 \\
Average     & 46.7 & 46.2 & 50.1 & 38.5 & 38.0 & 41.6 & 39.7 & 39.2 & 42.0 \\ \hline \hline
\end{tabular}
}
\end{center}
\caption{Evaluation results on GPT-4V, LLaVA-NeXT-34B, and InterLM-XComposer2-4KHD-7B~(InternLM-X2-4K-7B for short) under Q+I setting in term of different instruction format.}
\label{tab:3model_instruct}
\end{table*}
In Table~\ref{tab:3model_compare}, we present the performance of three representative MLLMs, \emph{i.e.}, GPT-4V, LLaVA-NeXT-34B, and InternLM-XComposer2-4KHD-7B, across different evaluation settings in multimodal financial contexts. Despite the highest performance of GPT-4V at setting-Q, the inclusion of chart information in setting-Q+I significantly improves performance, illustrating its strong capability to integrate visual content. Conversely, InternLM-XComposer2-4KHD-7B also demonstrates significant improvements, attributed to its higher resolution for visual processing. In contrast, LLaVA-NeXT-34B shows minimal gains, suggesting the need to enhance its CLIP-based visual alignment for charts.
Additionally, all MLLMs improve when captions is added to questions (from setting-Q to setting-Q+C), but starting from setting-Q+I, only LLaVA-NeXT-34B and InternLM-XComposer2-4KHD-7B could further benefit from the addition of captions (setting-Q+I+C), while GPT-4V experiences a decrease. This indicates that captions generated by GPT-4V can provide additional useful information to other MLLMs. However, since the captions only contain part of the information from the charts, they naturally struggle to offer more information to GPT-4V itself.

Overall, MLLMs still show limited capabilities in integrating visual content in financial applications. Supplementing effective textual information can also significantly aid MLLMs in achieving better performance. These findings highlight the importance of fine-tuning multimodal architectures and optimizing caption integration methods to bridge the gap between visual and textual.

\subsection{Ablation Study}

\subsubsection{The Effect of Instruction}

To further explore the MLLMs to different instructions, we design two extra types of instructions and display the results in Table~\ref{tab:3model_instruct}~(Detailed instructions can be found in appendix). 
The first type, \textbf{Simple},  simplifies existing instructions by directly requiring models to answer questions based on provided charts, removing any additional guidance or contextual cues to test the models' ability to perform with minimal input.  
The second type, \textbf{ICL}, adds three sets of question-reason-answer combinations as prompts to study the in-context learning ability of MLLMs, enabling models to leverage contextual examples for better reasoning. We only include question-reason-answer pairs in few-shot instructions, as some MLLMs do not support multi-image input, which could otherwise complicate the evaluation process. This table reveals that simplified instructions lead to a slight decrease in model performance, about 1\%. This indicates that without additional information or example templates, the benefits of detailed instructions are limited. Conversely, the ICL results demonstrate that models achieves 3\%-4\% improvement from the original performance. Although image is not included in instructions, the improvement still suggests potential for enhancing
current instruction design. It also highlights significant opportunities for improvement through more innovative instruction design and comprehensive model training.

\subsubsection{The Effect of Language}

To further verify the impact of language on model performance, we translate all instructions, questions, and answers into English using GPT-4 and conduct experiments based on the translation. The experimental results are shown in Table~\ref{tab:3model_language}. From this table, we can find that LLaVA-NeXT-34B achieves an improvement of 1\%-3\% with English input, likely due to its stronger English proficiency compared to Chinese. In contrast, GPT-4V and InternLM-XComposer2-4KHD do not show significant changes, indicating their robustness in different languages. This suggests that language itself does not significantly affect the difficulty of answering questions, as the core challenge lies in the complexity of the financial and multimodal tasks rather than linguistic barriers. However, the results also highlight that LLaVA-NeXT-34B performs poorly in Chinese, which significantly affects its performance in CFBenchmark-MM, underscoring the importance of language-specific optimization for models targeting non-English domains. These findings emphasize the need for future research to address language-specific limitations and enhance multilingual capabilities in existing MLLMs, particularly for applications in diverse linguistic and cultural contexts. 
Overall, our findings reinforce that while language translation offers benefits to some MLLMs, the core difficulty lies in the complexity of the multimodal financial tasks.

\begin{table*}[ht]
\begin{center}
\resizebox{\textwidth}{!}{
\setlength\tabcolsep{25pt}
\begin{tabular}{c|cc|cc|cc}
\hline \hline
\multirow{2}{*}{Question} & \multicolumn{2}{c|}{GPT-4V} & \multicolumn{2}{c|}{LLaVA-NeXT-34B} & \multicolumn{2}{c}{InternLM-X2-4K-7B} \\ \cline{2-7} 
 & zh & en & zh & en & zh & en \\ \hline
Arithmetic  & 46.0 & 45.5 & 39.2 & 40.6 & 38.8 & 39.3  \\
Statistical & 52.1 & 53.8 & 42.8 & 45.9 & 44.6 & 42.8  \\
Structural  & 58.1 & 57.2 & 50.3 & 53.8 & 52.1 & 49.8  \\
Explanation & 24.8 & 25.2 & 15.8 & 16.1 & 27.7 & 26.5  \\
Knowledge   & 52.4 & 50.7 & 44.2 & 45.5 & 35.5 & 36.7  \\
Average     & 46.7 & 46.5 & 38.5 & 40.4 & 39.7 & 39.0  \\ \hline \hline
\end{tabular}
}
\end{center}
\caption{Evaluation results on GPT-4V, LLaVA-NeXT-34B, and InterLM-XComposer2-4KHD-7B~(InternLM-X2-4K-7B for short) under Q+I setting in term of different instruction format.}
\label{tab:3model_language}
\end{table*}

\begin{figure*}[ht]
    \centering
    \includegraphics[width=1\linewidth]{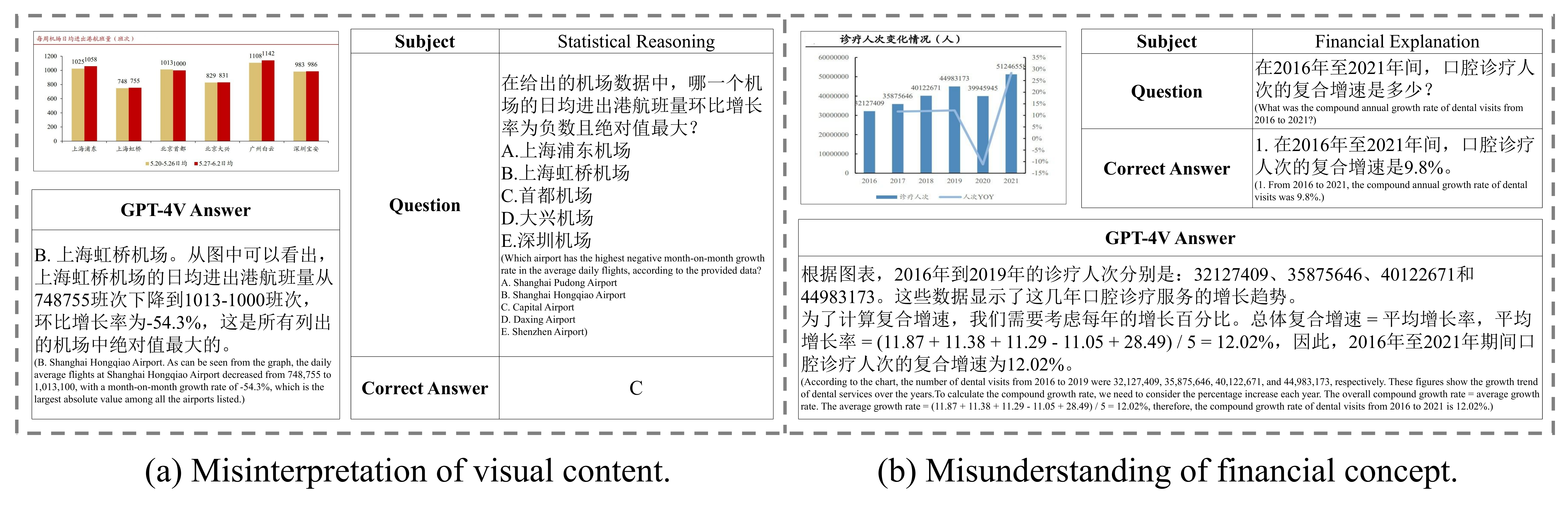}
    \caption{Failure cases in CFBenchmark-MM by GPT-4V with reasons and answers.}
    \label{fig:case_figure}
\end{figure*}

\subsection{Case Study}
To further explore issues with MLLMs in multimodal financial contexts, we present two typical cases of incorrect responses from GPT-4V in Figure~\ref{fig:case_figure} and analyze the reasons to these errors.
In Figure~\ref{fig:case_figure}(a), we showcase an example from statistical reasoning, where the response from GPT-4V confounds the traffic data from different periods and airports, failing to extract the correct visual information to answer the question. Furthermore, in the response, the calculation of GPT-4V is also incorrect, indicating that its mathematical abilities need further improvement, particularly in handling complex numerical data.
In Figure\ref{fig:case_figure}~(b), we present an example from financial explanation, where the response from GPT-4V reveals the failure in distinguishing ``compound growth rate'' and ``average growth rate'' from the question and chart, highlighting an incomplete understanding of GPT-4V in financial concepts. Moreover, when we query the relationship between these two concepts in textual form with GPT-4V, it correctly identifies their relationship, which shows that aligning visual-textual concepts poses more challenges than textual-textual concepts. This discrepancy underscores the need for improved multimodal alignment in financial contexts, particularly in integrating visual data with textual analysis to enhance the accuracy and reliability of MLLMs in real-world financial applications. These findings suggest that while MLLMs have made significant strides, there is still considerable room for improvement in their ability to process and interpret complex multimodal financial data.
Additionally, our analysis reveals that the errors may stem from limitations in training data diversity and challenges in visual processing, which indicate a critical need for advanced fine-tuning strategies. These shortcomings illustrate that even state-of-the-art MLLMs require further enhancement to handle the intricacies of financial data analysis and multimodal integration.

\section{Conclusion}

This paper has explored the development and implication of Multimodal Large Language Models~(MLLMs) in multimodal finance application, highlighting the crucial role of integrating texts, charts and figures to improve decision-making accuracy and efficiency. 
Therefore, we introduce CFBenchmark-MM, a comprehensive financial benchmark with over 9,000 image-question pairs, including various data representations such as tables, line charts, pie charts, and structural diagrams. 
Our evaluation settings have assessed the capabilities of MLLMs in handling multimodal information, from textual data to visual content integration.
These findings indicate that although MLLMs are capable of possessing inherent financial knowledge, their efficiency in processing multimodal information is still limited, highlighting the need for further improvements in the multimodal context. 
Our research underscores the significant yet underexploited potential of MLLMs in financial analysis and advocates for their enhanced application in the financial domain. 
The insights call for an ongoing refinement of the MLLMs to better utilize multimodal information, with the aim of achieving more robust and accurate financial decision making.

\bibliographystyle{ACM-Reference-Format}
\bibliography{cite}

\clearpage

\appendix

\section{Limitation and Future Work}~\label{sec:limit}

In this paper, we investigate the evaluation of Multimodal Large Language Models (MLLMs) in financial application and propose a benchmark and an evaluation framework, CFBenchmark-MM. 
The benchmark segregates into five question categories, analyzing 12 open-source MLLMs and 1 close-source MLLM, GPT-4V. 
The benchmark is scheduled for release following the publication of this paper.
However, our evaluations predominantly concentrate on open-source MLLMs, with a noticeable absence of testing on certain proprietary models such as Claude and Gemini. Additionally, while our five categories of questions encompass a broad range of real-world financial applications, they do not extend into specialized sub-disciplines within finance, such as investments and banking.
In constructing our benchmark, we effectively utilize multimodal data from financial research reports to generate questions and answers, primarily focusing on Chinese-language reports. 
Although an English version is provided, the exploration of multilingual benchmark remains constrained.

In the future, the research directions can be discussed from the following two perspective:

\begin{itemize}
    \item \textbf{Benchmark in finance}:
    1) Enhancing the task classification by breaking down the current five question types into more detailed financial sub-domains;
    2) Extending the linguistic scope of benchmark system by including additional languages such as French, Japanese, and German, moving beyond the current bilingual (Chinese-English) framework;
    3) Incorporating considerations of security and robustness in the design of future benchmarks to facilitate safer and more reliable applications of MLLMs in financial domain.
    \item \textbf{MLLMs in finance}:
    1) Expanding the application of MLLMs within the financial domain by leveraging more domain-specific multimodal data to optimize their performance;
    2) Developing specialized visual alignment modules as current reliance on pre-trained models like CLIP and ViT offers limited functionality in accurately representing financial charts;
    3) Investigating the potential of multi-agent systems in multimodal financial settings to understand how collaborative interactions among different agents can enhance model effectiveness and decision-making.
\end{itemize}

These proposed enhancements aim to deepen the integration, utility, and adaptability of MLLMs in financial domain, thereby augmenting their applicability in sophisticated financial analyses.

\section{Detailed Instructions and Prompts}~\label{sec:instruction}

In this section, we detail the instructions and prompts employed during the benchmark construction and evaluation. 
For the benchmark construction, we provide human annotators with textual instructions to ensure the quality of the benchmark during the multi-modal data collection~(Section~\ref{sec:instruction:collection}) and manual examination~(Section~\ref{sec:instruction:mannul}) stages. 
Additionally, corresponding to five question type, we design five different annotation prompts~(Section~\ref{sec:instruction:annotation}) for GPT-4V to convert financial analyses into corresponding questions and answers.
For the evaluation of the benchmark, we devise three types of prompts~(Section~\ref{sec:instruction:evaluation}) for MLLMs to answer the questions, namely, \textbf{original}, \textbf{simple}, and \textbf{ICL}. 
These prompts are structured to assess the ability of MLLMs to process and interpret the information under varying levels of complexity and context-specific guidance.

\subsection{Instructions for Multi-modal Data Collection}~\label{sec:instruction:collection}

\begin{tcolorbox}[breakable,auto outer arc, boxrule=0.5pt, colback=white!90!yellow, left=1mm, right=1mm, top=1mm, bottom=1mm]
\fontsize{8pt}{9pt}\selectfont
\begin{CJK}{UTF8}{gbsn}
本文档旨在指导标注人员准确收集和标注金融和统计类图表及其相关文字说明的数据，以构建一个高质量的金融图片及描述性文本数据集。为了可以准确的反映模型在实际金融场景下的建模能力，我们所有的金融多模态数据均来自于真实的研究报告和公告。\\
标注标准：
\begin{enumerate}
\item 仅选择清晰且质量高的金融或统计类图表，确保每个图表都伴随其相应的文字说明。
\item 文字说明应位于图表的前后文中，并详细阐明图表的内容和结论（超过50个汉字） 。
\item 对于统计类图表（如柱状图、折线图、饼图等） ，图表必须包含完整的数据标签和比例。确保图表的横纵坐标清晰标注，并且所有图形元素都易于辨识。 
\item 收集的表格必须清晰可读，表中的数据和文字信息应完整无缺。表头应被包含在截图中，以确保数据的完整性和上下文的清晰。 
\item 确保数据的准确性和一致性 。文字说明中提及的任何具体数据或比例，应在图表中有直接的对应或能够通过图表验证 。
\item 图片类型一共有五大类，分别是柱状图、折线图、表格、结构化图表以及其他类型的图表，除结构化图表外，标注过程中需确保各类数量均衡，其中结构化图表由于总体数量较少可以不受数量限制。
\end{enumerate}
数据存储标准：
\begin{enumerate}
\item 所有图像文件应保存为高分辨率的 PNG 格式 ，文件命名为\{number\}.png。
\item 描述性文本及其他标注信息以表格形式储存在 CSV 格式的文件中，表头分别为 
 file\_name,  description, figure\_type, date。
\end{enumerate}
\end{CJK}
(This document is designed to guide annotators in accurately collecting and annotating data related to financial and statistical charts along with their corresponding textual descriptions, in order to build a high-quality dataset of financial images and descriptive texts. To accurately reflect the model's ability in real financial scenarios, all our multimodal financial data come from actual research reports and announcements.
Annotating Standards:
\begin{enumerate}
    \item Only select clear and high-quality financial or statistical charts, ensuring each chart is accompanied by its corresponding textual explanation.
    \item Textual explanations should be placed in the context around the chart and detail the content and conclusions of the chart (over 50 Chinese characters).
    \item For statistical charts (such as histogram charts, line charts, pie charts, etc.), charts must include complete data labels and scales. Ensure the axes are clearly labeled and all graphical elements are easily identifiable.
    \item Collected tables must be clear and readable, with complete data and textual information. The header should be included in the screenshot to ensure the completeness of the data and clarity of the context.
    \item Ensure the accuracy and consistency of the data. Any specific data or ratios mentioned in the textual explanations should have a direct correspondence in the chart or be verifiable through the chart.
    \item There are five major types of images: histogram charts, line charts, tables, structured charts, and other types of charts. Except for structured charts, which are generally fewer in number and thus not subject to quantity limits, ensure a balanced distribution of other types during the annotating process.
\end{enumerate}
Data Storage Standards:
\begin{enumerate}
    \item All image files should be saved in high-resolution PNG format, with the file name as \{number\}.png.
    \item Descriptive texts and other labeling information should be stored in a CSV format file in table form, with headers as file\_name, description, figure\_type, date.)
\end{enumerate}
\normalsize
\end{tcolorbox}
\subsection{Prompts for Benchmark Annotation}~\label{sec:instruction:annotation}

\subsubsection{Arithmetic Reasoning}

\begin{tcolorbox}[breakable,auto outer arc, boxrule=0.5pt, colback=white!90!yellow, left=1mm, right=1mm, top=1mm, bottom=1mm]
\fontsize{8pt}{9pt}\selectfont
\begin{CJK}{UTF8}{gbsn}
以下是一份关于金融图表的描述性材料，请先判断该材料是否适合生成数值计算类问题。如果不适合，给出回答：“无法生成问题”；如果适合，请根据材料生成两个与材料相关的数值计算类的不定项选择题。每个问题应涵盖不同的统计概念，如最大值、最小值、比较、趋势分析、平均值计算或百分比计算等。问题和选项必须基于材料中的数据，并且确保每个选项有差异度，以避免生成过于接近的数值选项，增加题目的辨识度和挑战性。请随机设置三到五个选项，避免每个问题选项数量固定。每个问题的答案应能直接从提供的数据中分析得出。生成问题的格式为：问题1：\{question1\}\textbackslash nA.\{choice content\}\textbackslash nB.\{choice content\}\textbackslash n...\textbackslash n\textbackslash n
问题2：\{question2\}\textbackslash nA.\{choice content\}\textbackslash nB.\{choice content\}\textbackslash n...\textbackslash n\textbackslash n描述性材料为：\{caption\}
\end{CJK}\\
(Here is a descriptive document about financial charts. First, assess whether this material is suitable for generating \textbf{Arithmetic Reasoning} problems. If it is not suitable, respond with: 'Unable to generate questions'. If it is suitable, create two \textbf{Arithmetic Reasoning} multiple-choice problems related to the document. Each question should cover different statistical concepts such as maximum, minimum, comparison, trend analysis, average calculation, or percentage calculation. The questions and options must be based on the data in the document, and ensure each option is distinct to avoid generating overly similar numerical options, increasing the distinctiveness and challenge of the questions. Set three to five options randomly to avoid a fixed number of options for each question. The answer to each question should be directly analyzable from the provided data.
Format of the questions:  Question1:\{question1\}\textbackslash n
A.\{choice content\}\textbackslash n
B.\{choice content\}\textbackslash n
...\textbackslash n\textbackslash n
Question2:\{question2\}\textbackslash n
A.\{choice content\}\textbackslash n
B.\{choice content\}\textbackslash n
...\textbackslash n\textbackslash n
Descriptive material: \{caption\} )

\normalsize
\end{tcolorbox}

\subsubsection{Statistical Reasoning}

\begin{tcolorbox}[breakable,auto outer arc, boxrule=0.5pt, colback=white!90!yellow, left=1mm, right=1mm, top=1mm, bottom=1mm]
\fontsize{8pt}{9pt}\selectfont
\begin{CJK}{UTF8}{gbsn}
以下是一份关于金融图表的描述性材料，请先判断该材料是否适合生成数据统计类问题。如果不适合，给出回答：“无法生成问题”；如果适合，请根据材料生成两个与材料相关的数据统计类的不定项选择题（例如：xxx的数据是多少？）。问题和选项必须基于材料中的数据，并且确保每个选项有差异度，以避免生成过于接近的数值选项，增加题目的辨识度和挑战性。请随机设置三到五个选项，避免每个问题选项数量固定。每个问题的答案应能直接从提供的材料中找到。 
生成问题的格式为：问题1：\{question1\}\textbackslash n
A.\{choice content\}\textbackslash n
B.\{choice content\}\textbackslash n
...\textbackslash n\textbackslash n
问题2：\{question2\}\textbackslash n
A.\{choice content\}\textbackslash n
B.\{choice content\}\textbackslash n
...\textbackslash n\textbackslash n
描述性材料为：\{caption\}
\end{CJK}\\
(Here is a descriptive document about financial charts. First, assess whether this material is suitable for generating \textbf{Statistical Reasoning} questions. If it is not suitable, respond with: 'Unable to generate questions'. If it is suitable, create two multiple-choice \textbf{Statistical Reasoning} questions related to the document (for example: What is the data for xxx?). The questions and options must be based on the data in the document, and ensure each option is distinct to avoid generating overly similar numerical options, increasing the distinctiveness and challenge of the questions. Set three to five options randomly to avoid a fixed number of options for each question. The answer to each question should be directly found in the provided material. Format of the questions:  Question1:\{question1\}\textbackslash n
A.\{choice content\}\textbackslash n
B.\{choice content\}\textbackslash n
...\textbackslash n\textbackslash n
Question2:\{question2\}\textbackslash n
A.\{choice content\}\textbackslash n
B.\{choice content\}\textbackslash n
...\textbackslash n\textbackslash n
Descriptive material: \{caption\} )
\normalsize
\end{tcolorbox}

\subsubsection{Structural Reasoning}

\begin{tcolorbox}[breakable,auto outer arc, boxrule=0.5pt, colback=white!90!yellow, left=1mm, right=1mm, top=1mm, bottom=1mm]
\fontsize{8pt}{9pt}\selectfont
\begin{CJK}{UTF8}{gbsn}
以下是一份关于结构化图表的描述性材料，这一类图表以文字为核心，往往阐述了某一流程或某种结构。请先判断该材料是否适合生成结构推理类问题。如果不适合，给出回答：“无法生成问题”；如果适合，请根据材料生成两个与材料相关的结构推理类的不定项选择题。问题和选项必须基于材料中的文字，聚焦在流程图和结构图的识别与分析。请随机设置三到五个选项，避免每个问题选项数量固定。每个问题的答案应能直接从提供的材料中找到。 
生成问题的格式为：问题1：\{question1\}\textbackslash n
A.\{choice content\}\textbackslash n
B.\{choice content\}\textbackslash n
...\textbackslash n\textbackslash n
问题2：\{question2\}\textbackslash n
A.\{choice content\}\textbackslash n
B.\{choice content\}\textbackslash n
...\textbackslash n\textbackslash n
描述性材料为：\{caption\}
\end{CJK}\\
(Here is a descriptive document about structured charts, which are primarily text-based and often explain a particular process or structure. First, assess whether this material is suitable for generating \textbf{Structural Reasoning} questions. If it is not suitable, respond with: 'Unable to generate questions'. If it is suitable, create two multiple-choice \textbf{structure reasoning} questions related to the document. The questions and options must be based on the text in the document, focusing on the identification and analysis of flowcharts and structure diagrams. Set three to five options randomly to avoid a fixed number of options for each question. The answer to each question should be directly found in the provided material.
Format of the questions:  Question1:\{question1\}\textbackslash n
A.\{choice content\}\textbackslash n
B.\{choice content\}\textbackslash n
...\textbackslash n\textbackslash n
Question2:\{question2\}\textbackslash n
A.\{choice content\}\textbackslash n
B.\{choice content\}\textbackslash n
...\textbackslash n\textbackslash n
Descriptive material: \{caption\} )

\normalsize
\end{tcolorbox}

\subsubsection{Financial Explanation}

\begin{tcolorbox}[breakable,auto outer arc, boxrule=0.5pt, colback=white!90!yellow, left=1mm, right=1mm, top=1mm, bottom=1mm]
\fontsize{8pt}{9pt}\selectfont
\begin{CJK}{UTF8}{gbsn}
以下是一份金融图表的描述性材料，请先判断该材料是否包含一定的金融或数据类信息，能否生成金融解读问题。如果不适合，给出回答：“无法生成问题”；如果适合，请根据材料生成两个与材料相关的金融解读类的问题。这些问题必须涉及到图表所展示的数据或其他信息，给出的金融解读问题可以是趋势变化或者数值比较等等与图表紧密相关的问题。问题需要具有一定的复杂性，但问题的答案必须能够直接从材料数据中分析得出，不需要对数据变化的深层次原因或意义进行探讨。 
生成问题的格式为：
问题1：\{question1\}\textbackslash n 
问题2：\{question2\}\textbackslash n
描述性材料为：\{caption\}
\end{CJK}\\
(Here is a descriptive document about a financial chart. First, assess whether the material contains sufficient financial or data-related information to generate \textbf{Financial Explanation} questions. If it is not suitable, respond with: 'Unable to generate questions'. If it is suitable, create two \textbf{Financial Explanation} questions related to the document. These questions must involve the data or other information displayed in the charts, and the financial interpretation questions could be about trend changes, numerical comparisons, or other issues closely related to the chart. The questions should be sufficiently complex, but the answers must be directly analyzable from the data provided in the material, without needing to explore the deeper reasons or implications of data changes.
Format of the questions:  Question1:\{question1\}\textbackslash n
Question2:\{question2\}\textbackslash n
Descriptive material: \{caption\} )
\normalsize
\end{tcolorbox}

\subsubsection{Financial Knowledge}
\begin{tcolorbox}[breakable,auto outer arc, boxrule=0.5pt, colback=white!90!yellow, left=1mm, right=1mm, top=1mm, bottom=1mm]
\fontsize{8pt}{9pt}\selectfont
\begin{CJK}{UTF8}{gbsn}
以下是一份金融图表的描述性材料，在这类材料中，往往包含了一些金融领域的专业术语。请先判断该材料是否包含一定的金融信息，能否生成金融知识类问题。如果不适合，给出回答：“无法生成问题”；如果适合，请根据材料生成两个与材料相关的金融知识类的问题，这些问题必须涉及到材料给出的金融术语，来考验答题者对这些专业名词的理解。 
生成问题的格式为：
问题1：\{question1\}\textbackslash n 
问题2：\{question2\}\textbackslash n
描述性材料为：\{caption\}
\end{CJK}\\
(Here is a descriptive document about a financial chart, which often includes some professional financial terminology. First, assess whether the material contains sufficient financial information to generate  \textbf{Financial Knowledge} questions. If it is not suitable, respond with: 'Unable to generate questions'. If it is suitable, create two  \textbf{Financial Knowledge} questions related to the document. These questions must involve the financial terminology provided in the material, to test the respondents' understanding of these professional terms. 
Format of the questions:  Question1:\{question1\}\textbackslash n
Question2:\{question2\}\textbackslash n
Descriptive material: \{caption\} )
\normalsize
\end{tcolorbox}

\subsection{Instructions for Manual Examination}~\label{sec:instruction:mannul}
\begin{tcolorbox}[breakable,auto outer arc, boxrule=0.5pt, colback=white!90!yellow, left=1mm, right=1mm, top=1mm, bottom=1mm]
\fontsize{8pt}{9pt}\selectfont
\begin{CJK}{UTF8}{gbsn}
1. 问题的验证： 我们首先验证所提出的问题是否完全可以使用所提供的多模态信息来回答。这包括检查问题是否直接关联到给定的文本、图表或图像内容，以确保所有问题都能够凭借所给信息找到答案。 \\
2. 客观问题的格式与准确性检查： 对于客观类型的问题，我们细致检查选项的格式是否正确，并确保所有提供的答案选项中包含一个准确且无歧义的正确答案。这一步骤确保选择题的每个选项都清晰、合适，并且正确答案在选项中明确标出。 \\
3. 主观问题的答案验证： 对于主观问题，我们确认答案的准确性，并要求答案以要点格式呈现。这意味着答案应简洁明了，直接点出关键信息，并确保这些信息能够准确反映问题所需的内容。 \\
4. 标记与图表无关或信息不足的问题： 我们将手动标记那些与图表无直接关联或无法从图表中得到正确信息的问题。这包括识别和标记那些题干与图表内容不匹配或图表无法提供足够信息以回答问题的情况。 \\
\end{CJK}
(1. Question Verification: We first verify whether the questions posed can be entirely answered using the provided multimodal information. This includes checking if the questions are directly related to the given text, charts, or image content, to ensure that all questions can be answered based solely on the provided information.\\
2. Format and Accuracy Check for Objective Questions: For objective-type questions, we meticulously examine whether the format of the options is correct and ensure that among all the provided answer options, there is one accurate and unambiguous correct answer. This step ensures that each option in multiple-choice questions is clear, appropriate, and the correct answer is distinctly marked within the options.\\
3. Verification of Answers for Subjective Questions: For subjective questions, we confirm the accuracy of answers and require that the answers be presented in a bullet-point format. This means the answers should be concise and directly highlighting the key information and ensuring the information accurately reflects the content to address the question. \\
4. Marking Questions Irrelevant to Charts or Lacking Information: We mark questions that have no direct connection to the charts or cannot obtain information from the charts. This includes identifying and marking instances where the question does not match the content of the chart or the chart fails to provide sufficient information to answer the question.)
\normalsize
\end{tcolorbox}

\subsection{Prompts Benchmark Evaluation}~\label{sec:instruction:evaluation}

\subsubsection{Original}
\begin{tcolorbox}[breakable,auto outer arc, boxrule=0.5pt, colback=white!90!yellow, left=1mm, right=1mm, top=1mm, bottom=1mm]
\fontsize{8pt}{9pt}\selectfont
\begin{CJK}{UTF8}{gbsn}
以下是一道关于\{subject\}的题目，一张金融场景中的图表和对应的文字描述，请基于这张图表和文字描述回答问题。图表和文字描述可能为空，如果其为空时，请忽略其内容。这道题可能是多项选择题，也可能是简答题。针对多项选择题，请你选出所有的正确选项；针对简答题，请你逐点回答这个问题。这张图表可能是表格、结构图、饼图、折线图或柱状图。请在回答时仔细浏览图表的每个组成部分，包括但不限于：标题、X轴和Y轴的标签、图例说明、各个数据系列的颜色和形状、图表中显示的数据点和数值。如果是表格，请详细浏览的标题和内容。如果是饼图，请说明每个扇区的百分比和扇区名称。如果数值没有标注在图上，请你根据X轴和Y轴判断数值大小，并基于判断的数值大小进行回答。题目为\{question\}。图表为\{<IMAGE>\}。图表描述为\{caption\}。
\end{CJK}\\
(Here is a question about \{subject\}, featuring a chart from financial domain and its corresponding textual description. Based on the chart and text provided, please answer the question. The chart and text may be absent; if so, ignore their content. The question may be multiple-choice or a short-answer type. For multiple-choice questions, select all correct options; for short-answer questions, respond to each point in the question. The chart could be a table, structural diagram, pie chart, line chart, or bar chart. When answering, carefully examine each component of the chart, including but not limited to: the title, labels on the X and Y axes, the legend, the colors and shapes of data series, and the data points and values displayed. If it is a table, closely review the title and content. If it is pie charts, specify the percentage and name of each sector. If values are not marked on chart, determine the size of the values based on the X, Y axes and answer based on these estimated values.
The question is \{question\}. The chart is \{<IMAGE>\}. The chart description is \{caption\}.)
\normalsize
\end{tcolorbox}

\subsubsection{Simple}
\begin{tcolorbox}[breakable,auto outer arc, boxrule=0.5pt, colback=white!90!yellow, left=1mm, right=1mm, top=1mm, bottom=1mm]
\fontsize{8pt}{9pt}\selectfont
\begin{CJK}{UTF8}{gbsn}
以下是一道关于\{subject\}的题目，一张金融场景中的图表和对应的文字描述，请基于这张图表和文字描述回答问题。图表和文字描述可能为空，如果其为空时，请忽略其内容。这道题可能是多项选择题，也可能是简答题。针对多项选择题，请你选出所有的正确选项；针对简单题，请你逐点回答这个问题。题目为\{question\}。图表为\{<IMAGE>\}。图表描述为\{caption\}。
\end{CJK}\\
(Here is a question about \{subject\}, featuring a chart from financial domain and its corresponding textual description. Based on the chart and text provided, please answer the question. The chart and text may be absent; if so, ignore their content. The question may be multiple-choice or a short-answer type. For multiple-choice questions, select all correct options; for short-answer questions, respond to each point in the question. 
The question is \{question\}. The chart is \{<IMAGE>\}. The chart description is \{caption\}.)
\normalsize
\end{tcolorbox}
\subsubsection{ICL}

\textbf{Arithmetic Reasoning}
\begin{tcolorbox}[breakable,auto outer arc, boxrule=0.5pt, colback=white!90!yellow, left=1mm, right=1mm, top=1mm, bottom=1mm, breakable]
\fontsize{8pt}{9pt}\selectfont
\begin{CJK}{UTF8}{gbsn}
案例1: 问题：从22年公司期间到23Q1期间，期间费率的增长是多少个百分点？\textbackslash n
A. 10.46\%\textbackslash n
B. 12.46\%\textbackslash n
C. 13.46\%\textbackslash n
D. 22.46\%\textbackslash n
E. 25.70\% \textbackslash n
回答：从图中可以看出，22年公司期间费率44.16\%，同增13.1pct，其中销售费率38.63\%，同增12.3pct;23Q1期间费率56.62\%，同增25.7pct，其中销售费率增长21.2pct，销售费用增长主要系门店数量和销售人员增加，工资薪金、市场推广费、门店租赁相关支出等增加所致，所以答案选择：E\textbackslash n \textbackslash n \textbackslash n 

案例2: 问题：从2015年到2022年，公司的管理费用率下降了多少个百分点？
A.8.0\%\textbackslash n
B.8.1\%\textbackslash n
C.8.2\%\textbackslash n
D.8.3\%\textbackslash n
E.8.4\%\textbackslash n
回答：从图中可以看出，从公司收入保持较快增长，规模效应释放下，费用率持续改善，管理费用率由2015年的10.9\%下滑至2022年的2.6\%，销售费用率由2015年的2.9\%下滑至2022年的1.1\%，财务费用率由2017年的0.8\%下滑至2022年的0.5\%，所以答案选择：D\textbackslash n \textbackslash n \textbackslash n \textbackslash n 

案例3: 问题：从7月13日至8月13日，理财子产品的单位净值环比下降了多少个百分点？
A.0.28\%\textbackslash n
B.0.50\%\textbackslash n
C.0.58\%\textbackslash n
D.2.00\%\textbackslash n
E.2.61\%\textbackslash n
回答：从图中可以看出，净值变化来看，截至8月13日，理财子产品单位净值为1.036，环比下降0.58\%;累计净值为1.039，环比下降2.61\%，所以答案选择：C\textbackslash n \textbackslash n \textbackslash n \textbackslash n 

我将提供给你一张图片和一个算数题，请仿照以上案例的问答结构，请依据该图片选出其中的正确答案。请你一步步思考和推理，先描述分析图片，再解释原因，最后给出选项。题目为\{question\}。图表为\{<IMAGE>\}。
\end{CJK}\\
(Case 1: Question: What is the percentage point increase in the expense ratio from 2022 to Q1 2023? \textbackslash n
A. 10.46\%\textbackslash n
B. 12.46\%\textbackslash n
C. 13.46\%\textbackslash n
D. 22.46\%\textbackslash n
E. 25.70\% \textbackslash n
Answer: The chart shows that the company's expense ratio was 44.16\% in 2022, with a year-on-year increase of 13.1pct, including a sales expense ratio of 38.63\%, which increased by 12.3pct; the Q1 2023 expense ratio was 56.62\%, with a year-on-year increase of 25.7pct, where the sales expense ratio increased by 21.2pct. The increase in sales expenses was mainly due to an increase in the number of stores and sales personnel, as well as increases in wages, marketing expenses, store leases, and other expenditures. Therefore, the answer is: E\textbackslash n \textbackslash n \textbackslash n 

Case 2: Question: From 2015 to 2022, how many percentage points did the company's management expense ratio decrease?
A.8.0\%\textbackslash n
B.8.1\%\textbackslash n
C.8.2\%\textbackslash n
D.8.3\%\textbackslash n
E.8.4\%\textbackslash n
Answer: The chart shows that as the company's revenue maintained rapid growth and scale effects were realized, the expense ratio continued to improve. The management expense ratio decreased from 10.9\% in 2015 to 2.6\% in 2022, the sales expense ratio from 2.9\% in 2015 to 1.1\% in 2022, and the financial expense ratio from 0.8\% in 2017 to 0.5\% in 2022. Thus, the answer is: D\textbackslash n \textbackslash n \textbackslash n \textbackslash n 

Case 3: Question: From July 13 to August 13, how many percentage points did the net value per unit of the wealth management sub-product decrease?
A.0.28\%\textbackslash n
B.0.50\%\textbackslash n
C.0.58\%\textbackslash n
D.2.00\%\textbackslash n
E.2.61\%\textbackslash n
Answer: The chart shows that as of August 13, the net value per unit of the wealth management sub-product was 1.036, a month-on-month decrease of 0.58\%; the cumulative net value was 1.039, a month-on-month decrease of 2.61\%. Therefore, the answer is: C\textbackslash n \textbackslash n \textbackslash n \textbackslash n 

I will provide you with an image and a math question. Please follow the structure of the examples above to select the correct answer based on the image. Think and reason step by step, first analyze the image, then explain the reasons, and finally give the option. The question is \{question\}. The chart is \{<IMAGE>\}.)
\normalsize
\end{tcolorbox}

\textbf{Statistical Reasoning}

\begin{tcolorbox}[breakable,auto outer arc, boxrule=0.5pt, colback=white!90!yellow, left=1mm, right=1mm, top=1mm, bottom=1mm, breakable]
\fontsize{8pt}{9pt}\selectfont
\begin{CJK}{UTF8}{gbsn}
案例1: 问题：本年首日涨幅最大的企业是哪家？\textbackslash n
A.巨能股份\textbackslash n
B.天力复合\textbackslash n
C.晟楠科技\textbackslash n
D.华信永道\textbackslash n
E.以上都不是\textbackslash n
回答：从图中可以看出，本年首日涨幅最大的前3家企业由巨能股份(+227.27\%)、天力复合(+92.83\%)、晟楠科技(+84.64\%)保持。本周天力复合和华信永道上市，首日收盘保持正增长，所以答案选择：A\textbackslash n \textbackslash n \textbackslash n 

案例2: 问题：2023年2月至4月期间，城投拿地占比的趋势是怎样的？\textbackslash n
A.持续上升\textbackslash n
B.持续下滑\textbackslash n
C.先上升后下滑\textbackslash n
D.先下滑后上升\textbackslash n
E.保持不变\textbackslash n
回答：从图中可以看出，2022年7月-2023年8月单月情况来看，2022年11、12、1月城投企业拿地额比例均超过了50\%，分别为57\%、67\%、54\%;而今年2月-4月城投拿地占比持续下滑，5月-8月城投拿地占比有所提升，维持低位，所以答案选择：B\textbackslash n \textbackslash n \textbackslash n \textbackslash n 

案例3: 问题：本周港股哪个行业的涨幅最高？\textbackslash n
A.资讯科技业\textbackslash n
B.可选消费\textbackslash n
C.工业\textbackslash n
D.公共事业\textbackslash n
E.电讯业\textbackslash n
回答：从图中可以看出，本周港股多数行业收涨。n其中资讯科技业、可选消费、工业本周涨幅居前，分别达到+7.9\%/+6.0\%/+5.1\%。n本周仅有公共事业、电讯业、能源业收跌，分别录得-0.1\%/-2.6\%/-4.0\%，所以答案选择：A\textbackslash n \textbackslash n \textbackslash n \textbackslash n 

我将提供给你一张图片和一个统计题，请仿照以上案例的问答结构，请依据该图片选出其中的正确答案。请你一步步思考和推理，先描述分析图片，再解释原因，最后给出选项。题目为\{question\}。图表为\{<IMAGE>\}。
\end{CJK}\\
(Case 1: Question: Which company had the highest increase on the first day of this year? \textbackslash n
A. JuNeng Shares\textbackslash n
B. TianLi Composite\textbackslash n
C. ShengNan Technology\textbackslash n
D. HuaXin YongDao\textbackslash n
E. None of the above\textbackslash n
Answer: The chart shows that the top three companies with the highest first-day gains this year are JuNeng Shares (+227.27\%), TianLi Composite (+92.83\%), and ShengNan Technology (+84.64\%). This week, TianLi Composite and HuaXin YongDao were listed and maintained positive growth at close, so the answer is: A\textbackslash n \textbackslash n \textbackslash n 

Case 2: Question: What was the trend of the urban investment land acquisition ratio from February to April 2023? \textbackslash n
A. Continuously rising\textbackslash n
B. Continuously falling\textbackslash n
C. Rising then falling\textbackslash n
D. Falling then rising\textbackslash n
E. Remained unchanged\textbackslash n
Answer: The chart shows that from July 2022 to August 2023 on a monthly basis, the proportion of land acquired by urban investment companies exceeded 50\% in November, December, and January, at 57\%, 67\%, and 54\% respectively. From February to April this year, the urban investment land acquisition ratio continuously declined, but from May to August, it increased slightly, remaining low. Therefore, the answer is: B\textbackslash n \textbackslash n \textbackslash n \textbackslash n 

Case 3: Question: Which industry in the Hong Kong stock market had the highest gain this week? \textbackslash n
A. Information Technology\textbackslash n
B. Consumer Discretionary\textbackslash n
C. Industrials\textbackslash n
D. Utilities\textbackslash n
E. Telecommunications\textbackslash n
Answer: The chart shows that most industries in the Hong Kong stock market ended the week higher. Information Technology, Consumer Discretionary, and Industrials led the gains, with increases of +7.9\%, +6.0\%, and +5.1\% respectively. This week, only Utilities, Telecommunications, and Energy sectors fell, recording declines of -0.1\%, -2.6\%, and -4.0\% respectively. Therefore, the answer is: A\textbackslash n \textbackslash n \textbackslash n \textbackslash n 

I will provide you with an image and a statistical question. Please follow the structure of the examples above to select the correct answer based on the image. Think and reason step by step, first analyze the image, then explain the reasons, and finally give the option. The question is \{question\}. The chart is \{<IMAGE>\}.)
\normalsize
\end{tcolorbox}

\textbf{Structural Reasoning}

\begin{tcolorbox}[breakable,auto outer arc, boxrule=0.5pt, colback=white!90!yellow, left=1mm, right=1mm, top=1mm, bottom=1mm, breakable]
\fontsize{8pt}{9pt}\selectfont
\begin{CJK}{UTF8}{gbsn}
案例1: 问题：首创证券是由哪家公司以整体变更方式设立的？\textbackslash n
A.首创经纪\textbackslash n
B.首创集团\textbackslash n
C.首创有限\textbackslash n
D.首创投资\textbackslash n
E.首创控股\textbackslash n
回答：首创证券系由首创有限于2020年9月以整体变更方式设立，并于2022年12月A股上市，是A股第42家上市证券公司。首创有限的前身为首创经纪于2000年2月设立，于2004年1月更名为首创有限，所以答案选择：C\textbackslash n \textbackslash n \textbackslash n 

案例2: 问题：佳木斯电机股份有限公司因为什么原因在2017年收到了证监会的市场禁入决定书和行政处罚书？\textbackslash n
A.少结转主营业务成本\textbackslash n
B.少计销售费用\textbackslash n
C.虚增2013、2014年利润\textbackslash n
D.延迟公布财务信息\textbackslash n
E.违规股票交易\textbackslash n
回答：从图中可以看出，公司电机研发历史超过80年。公司前身1937年成立于沈阳，1953年改名为佳木斯电厂，成功研制出我国第一台防爆电机，并于第一个五年计划开始我国防爆电机的大规模生产，1998年加入哈电集团，2000年公司改制成股份有限公司，佳木斯电机股份有限公司正式成立，2012年公司成功借壳上市，2017年证监会向佳电股份下发了市场禁入决定书和行政处罚书，主要原因：公司通过少结转主营业务成本、少计销售费用等方式虚增2013、2014年利润，2015年再将前期调节的利润进行消化，所以答案选择：A,B,C\textbackslash n \textbackslash n \textbackslash n \textbackslash n 

案例3: 问题：在电改的底层价格机制中，发电价格主要与哪种能源的价格联动？\textbackslash n
A.石油\textbackslash n
B.天然气\textbackslash n
C.煤炭\textbackslash n
D.太阳能\textbackslash n
E.风能\textbackslash n
回答：从图中可以看出，电改最终将形成一个稳定金字塔，底层的价格机制为【发电价格（煤电联动）+峰谷价差+辅助服务电价（待落地）】、中层的市场机制【现货及中长期电力交易市场+辅助服务市场】、顶层就是【电价恒等式，我们认为伴随中国电力结构的转型，过去以燃煤标杆电价为主的定价体系将逐步转化为电力的恒等式，即各类电源的成本应当殊途同归，比如绿电电价+辅助服务成本=火电电价+环境成本】，所以答案选择：C\textbackslash n \textbackslash n \textbackslash n \textbackslash n 

我将提供给你一张图片和一个结构分析题，请仿照以上案例的问答结构，请依据该图片选出其中的正确答案。请你一步步思考和推理，先描述分析图片，再解释原因，最后给出选项。题目为\{question\}。图表为\{<IMAGE>\}。
\end{CJK}\\
(Case 1: Question: Which company established Beijing Securities through a whole change method? \textbackslash n
A. Beijing Brokerage\textbackslash n
B. Beijing Group\textbackslash n
C. Beijing Limited\textbackslash n
D. Beijing Investment\textbackslash n
E. Beijing Holdings\textbackslash n
Answer: The chart shows that Beijing Securities was established by Beijing Limited in September 2020 through a whole change method and was listed on the A-share market in December 2022, becoming the 42nd listed securities company on the A-share market. Beijing Limited, originally known as Beijing Brokerage, was established in February 2000 and was renamed Beijing Limited in January 2004. Therefore, the answer is: C\textbackslash n \textbackslash n \textbackslash n

Case 2: Question: Why did Jiamusi Electric Machinery Co., Ltd. receive a market ban decision and administrative penalty from the CSRC in 2017? \textbackslash n
A. Underreported main business costs\textbackslash n
B. Underreported sales expenses\textbackslash n
C. Overstated profits for 2013 and 2014\textbackslash n
D. Delayed financial disclosure\textbackslash n
E. Illegal stock trading\textbackslash n
Answer: The chart indicates that the company, with over 80 years of R\&D history in motors, was initially established in Shenyang in 1937, renamed Jiamusi Electric Factory in 1953, produced China’s first explosion-proof motor, and started large-scale production of explosion-proof motors during the First Five-Year Plan. It joined Harbin Electric Group in 1998, restructured into a shareholding company in 2000, and Jiamusi Electric Machinery Co., Ltd. was officially established. It successfully listed through a reverse merger in 2012. In 2017, the CSRC issued a market ban decision and administrative penalty mainly because the company inflated its profits for 2013 and 2014 by underreporting main business costs and sales expenses, then adjusted the previously manipulated profits in 2015. Therefore, the answer is: A, B, C\textbackslash n \textbackslash n \textbackslash n \textbackslash n

Case 3: Question: In the underlying price mechanism of the electricity reform, with which energy source is the electricity price primarily linked? \textbackslash n
A. Oil\textbackslash n
B. Natural Gas\textbackslash n
C. Coal\textbackslash n
D. Solar Energy\textbackslash n
E. Wind Energy\textbackslash n
Answer: The chart illustrates that the electricity reform will form a stable pyramid structure, with the bottom layer's price mechanism being [electricity price (coal electricity linkage) + peak-valley price difference + auxiliary service electricity price (pending implementation)], the middle layer's market mechanism [spot and mid-long term electricity trading market + auxiliary services market], and the top layer being [electricity price equation, suggesting that as China’s power structure transforms, the pricing system primarily based on coal-fired benchmark electricity prices will gradually shift to an electricity price equation where the costs of various types of energy sources converge, such as green electricity price + auxiliary service costs = thermal electricity price + environmental costs]. Therefore, the answer is: C\textbackslash n \textbackslash n \textbackslash n \textbackslash n

I will provide you with an image and a structural analysis question. Please follow the structure of the examples above to select the correct answer based on the image. Think and reason step by step, first analyze the image, then explain the reasons, and finally give the option. The question is \{question\}. The chart is \{<IMAGE>\}.)
\normalsize
\end{tcolorbox}

\textbf{Financial Explanation}

\begin{tcolorbox}[breakable,auto outer arc, boxrule=0.5pt, colback=white!90!yellow, left=1mm, right=1mm, top=1mm, bottom=1mm, breakable]
\fontsize{8pt}{9pt}\selectfont
\begin{CJK}{UTF8}{gbsn}
案例1: 问题：在中国的电商发展历史中，哪个阶段B2B业务开始线上化，并且这一变化标志着什么？\textbackslash n
回答：1. 在中国的电商发展历史中，B2B业务在1998-2003阶段开始线上化。\textbackslash n
2. 这一变化标志着中国电子商务的起步和互联网的初步发展。B2B业务的线上化为企业间的交易提供了更高效、便捷的方式，推动了商业模式的创新和电子商务的进一步发展。这也为后续2C业务、O2O业务等电商模式的发展奠定了基础。\textbackslash n \textbackslash n \textbackslash n 

案例2: 问题：自2022年10月以来，商品住宅的当月销售均价表现出怎样的趋势？\textbackslash n
回答：1. 自2022年10月以来，商品住宅的当月销售均价出现了企稳回升的趋势。\textbackslash n
2. 这意味着在此之后，商品住宅的销售均价没有继续下降，而是开始出现回升的迹象。\textbackslash n
3. 特别是到了11月，商品住宅销售均价的同比增速脱离了负增长区间，进一步证实了企稳回升的趋势。\textbackslash n \textbackslash n \textbackslash n \textbackslash n 

案例3: 问题：在提到的新基建的7个重要领域中，哪些是与金融投资和技术创新紧密相关的？\textbackslash n
回答：1. 与金融投资紧密相关的新基建领域包括5G基建、城际高速铁路和城市轨道交通、新能源汽车充电桩、大数据中心、人工智能以及工业互联网。这些领域的建设和发展需要大量的资金投入，同时也为金融投资者提供了广阔的投资机会。\textbackslash n
2. 与技术创新紧密相关的新基建领域主要有5G基建、大数据中心、人工智能和工业互联网。这些领域都是当前科技发展的前沿，其建设和应用将推动技术创新和产业升级，为经济发展注入新的活力。\textbackslash n \textbackslash n \textbackslash n \textbackslash n 

我将提供给你一张图片和一个金融情景解释题，请仿照以上案例的问答结构，请依据该图片对于问题进行回答。请逐点对于题目进行回答。请尽可能全面的进行回答。请尽可能简洁的进行回答。题目为\{question\}。图表为\{<IMAGE>\}。
\end{CJK}\\
(Case 1: Question: In the history of e-commerce development in China, during which stage did B2B businesses begin to move online, and what did this change signify? \textbackslash n
Answer: 1. In the history of e-commerce development in China, B2B businesses started to go online during the 1998-2003 period.\textbackslash n
2. This change marked the beginning of e-commerce in China and the initial development of the internet. The online transition of B2B operations offered a more efficient and convenient way for business transactions, fostering innovation in business models and further development of e-commerce. It also laid the groundwork for the development of subsequent e-commerce models like B2C and O2O.\textbackslash n \textbackslash n \textbackslash n

Case 2: Question: Since October 2022, what trend has the average monthly selling price of residential properties shown?\textbackslash n
Answer: 1. Since October 2022, the average monthly selling price of residential properties has shown a trend of stabilization and recovery.\textbackslash n
2. This means that after that time, the average selling price of residential properties did not continue to decline but started to show signs of recovery.\textbackslash n
3. Particularly by November, the year-on-year growth rate of the average selling price for residential properties moved out of negative growth, further confirming the trend of stabilization and recovery.\textbackslash n \textbackslash n \textbackslash n \textbackslash n

Case 3: Question: Among the seven key areas of new infrastructure mentioned, which are closely related to financial investment and technological innovation?\textbackslash n
Answer: 1. The new infrastructure areas closely related to financial investment include 5G infrastructure, intercity high-speed rail and urban rail transit, new energy vehicle charging stations, big data centers, artificial intelligence, and the industrial internet. These areas require substantial capital for construction and development, also offering broad investment opportunities for financial investors.\textbackslash n
2. The areas closely related to technological innovation are mainly 5G infrastructure, big data centers, artificial intelligence, and the industrial internet. These sectors are at the forefront of current technological advancements, and their construction and application will drive technological innovation and industrial upgrading, injecting new vitality into economic development.\textbackslash n \textbackslash n \textbackslash n \textbackslash n

I will provide you with an image and a financial explanation question. Please follow the structure of the examples above to answer the question based on the image. Address the question point by point comprehensively and concisely. The question is \{question\}. The chart is \{<IMAGE>\}.)
\normalsize
\end{tcolorbox}

\textbf{Financial Knowledge}

\begin{tcolorbox}[breakable,auto outer arc, boxrule=0.5pt, colback=white!90!yellow, left=1mm, right=1mm, top=1mm, bottom=1mm, breakable]
\fontsize{8pt}{9pt}\selectfont
\begin{CJK}{UTF8}{gbsn}
案例1: 问题：销售费率同增的具体含义是什么，它反映了公司销售活动的哪些变化？\textbackslash n
回答：1. 销售费率同增的具体含义指的是与上一年同期相比，公司的销售费率出现了增长。在这个案例中，22年的销售费率同比增长了12.3个百分点，而23年第一季度的销售费率同比增长了21.2个百分点。\textbackslash n
2. 销售费率的增长反映了公司销售活动的多个方面的变化。首先，它可能意味着公司正在加大销售和市场推广的力度，这通常表现为销售费用（如工资薪金、市场推广费、门店租赁相关支出等）的增加。其次，销售费率的增长也可能反映了公司销售规模的扩大，例如门店数量的增加和销售人员的增多，这些都可能带来更高的销售费用。最后，销售费率的增长还可能意味着公司在销售效率方面存在挑战，需要更多的投入来维持或提高销售业绩。\textbackslash n \textbackslash n \textbackslash n 

案例2: 问题：单位净值和累计净值分别代表什么，它们在评估理财产品表现时有何重要性？\textbackslash n
回答：1. 单位净值代表每一份理财产品的净资产价值，是理财产品当前的总净资产除以发行的总份额得到的。它反映了每一份理财产品的实时价值，是投资者买卖理财产品的重要依据。\textbackslash n
2. 累计净值则代表了理财产品从成立以来的总收益情况，包括了所有的分红、拆分等因素。它是将历史的所有单位净值进行累加得到的，反映了理财产品的长期表现。\textbackslash n
3. 在评估理财产品表现时，单位净值和累计净值都非常重要。单位净值可以帮助投资者了解当前购买或赎回理财产品的价格，从而做出投资决策。而累计净值则可以帮助投资者了解理财产品的历史表现，评估其长期的投资价值。\textbackslash n
4. 同时观察单位净值和累计净值的变化，可以更全面地了解理财产品的表现。例如，单位净值上升可能意味着理财产品近期表现良好，而累计净值上升则意味着理财产品长期以来的总体表现不错。\textbackslash n \textbackslash n \textbackslash n \textbackslash n 

案例3: 问题：土地溢价率环比上周回落1.2个百分点是什么意思，上周土地溢价率如何计算？\textbackslash n
回答：1. ``环比上周回落1.2个百分点''的意思是相比上一周，这一周的土地溢价率下降了1.2个百分点。具体来说，``环比''是比较两个相邻周期的数据变化，这里是指与上一周相比；``回落''表示下降；``1.2个百分点''则是下降的具体数值，即下降了1.2\%的幅度。\textbackslash n
2. 计算回落的百分点，首先需要知道上周和本周的土地溢价率具体数值。假设上周土地溢价率为x\%，本周为y\%，那么环比回落的百分点就是上周的溢价率减去本周的溢价率，即(x - y)\%。根据题目给出的信息，全国土地溢价率本周为2\%，环比上周回落1.2个百分点，可以推算出上周全国土地溢价率为(2\% + 1.2\%) = 3.2\%。所以，计算方法是：上周土地溢价率 - 本周土地溢价率 = 回落的百分点。\textbackslash n \textbackslash n \textbackslash n \textbackslash n 

我将提供给你一张图片和一个金融知识拓展题，请仿照以上案例的问答结构，请依据该图片对于问题进行回答。请逐点对于题目进行回答。请尽可能全面的进行回答。请尽可能简洁的进行回答。题目为\{question\}。图表为\{<IMAGE>\}。
\end{CJK}\\
(Case 1: Question: What does the term ``year-on-year increase in the sales rate'' specifically mean, and what changes in the company's sales activities does it reflect? \textbackslash n
Answer: 1. ``Year-on-year increase in the sales rate'' refers to the growth in the company's sales rate compared to the same period last year. In this case, the sales rate in 2022 increased by 12.3 percentage points year-on-year, and in the first quarter of 2023, it increased by 21.2 percentage points.\textbackslash n
2. The growth in the sales rate reflects several aspects of changes in the company's sales activities. Firstly, it may indicate that the company is intensifying its sales and marketing efforts, often resulting in increased sales expenses such as wages, marketing costs, and store lease expenditures. Secondly, the increase in the sales rate could also reflect an expansion in the company's sales scale, such as an increase in the number of stores and sales personnel, which might lead to higher sales costs. Lastly, the increase in the sales rate could also signify challenges in sales efficiency, necessitating more input to maintain or improve sales performance.\textbackslash n \textbackslash n \textbackslash n

Case 2: Question: What do ``unit net value'' and ``cumulative net value'' represent, and why are they important in assessing the performance of financial products? \textbackslash n
Answer: 1. The ``unit net value'' represents the net asset value per share of a financial product, calculated by dividing the total net assets of the financial product by the total number of issued shares. It reflects the real-time value of each share of the financial product, crucial for investors when buying or selling.\textbackslash n
2. The ``cumulative net value'' represents the total returns of a financial product since its inception, including dividends, splits, etc. It is calculated by summing all historical unit net values, reflecting the long-term performance of the financial product.\textbackslash n
3. Both unit net value and cumulative net value are crucial in evaluating financial product performance. Unit net value helps investors understand the current price for purchasing or redeeming a financial product, thereby aiding investment decisions. Cumulative net value helps investors assess the historical performance of the product to gauge its long-term investment value.\textbackslash n
4. Observing changes in both unit net value and cumulative net value provides a more comprehensive understanding of a financial product's performance. For instance, an increase in unit net value might indicate good recent performance, while an increase in cumulative net value signifies good long-term overall performance.\textbackslash n \textbackslash n \textbackslash n \textbackslash n

Case 3: Question: What does a 1.2 percentage point decrease in the land premium rate compared to last week mean, and how was last week’s land premium rate calculated? \textbackslash n
Answer: 1. ``A decrease of 1.2 percentage points compared to last week'' means the land premium rate fell by 1.2 percentage points from last week. Specifically, ``compared to last week'' refers to the comparison between two consecutive periods; ``decrease'' indicates a decline; ``1.2 percentage points'' is the specific amount by which it decreased, i.e., a decline of 1.2\%.\textbackslash n
2. To calculate the percentage points of decline, one must know the specific land premium rates for last week and this week. If last week’s rate was x\% and this week’s is y\%, the week-over-week change is (x - y)\%. Based on the information provided, the national land premium rate this week is 2\%, having fallen 1.2 percentage points week-over-week, which implies last week’s rate was (2\% + 1.2\%) = 3.2\%. Thus, the calculation is: last week's land premium rate - this week's land premium rate = decline in percentage points.\textbackslash n \textbackslash n \textbackslash n \textbackslash n 

I will provide you with an image and a financial knowledge expansion question. Please follow the structure of the examples above to answer the question based on the image. Address the question point by point, aiming for comprehensive yet concise answers. The question is \{question\}. The chart is \{<IMAGE>\}.)
\normalsize
\end{tcolorbox}

\begin{table*}[ht]
\centering
\resizebox{\textwidth}{!}{
 \setlength\tabcolsep{10pt}
\begin{tabular}{c|cccc|cccc}
\hline\hline
\multirow{2}{*}{Question}    & \multicolumn{4}{c|}{Q}                   & \multicolumn{4}{c}{Q+C}                  \\ \cline{2-9} 
                             & InternLM2 & IntermLM2-X & Qwen & Qwen-VL & InternLM2 & IntermLM2-X & Qwen & Qwen-VL \\ \hline
\multirow{2}{*}{Arithmetic}  & 26.8      & 24.4        & 29.5 & 27.3    & 35.8      & 32.8        & 34.7 & 31.1    \\
                             & -         & -2.4        & -    & -2.2    & -         & -3.0        & -    & -3.6    \\
\multirow{2}{*}{Statistical} & 30.4      & 27.4        & 33.7 & 31.1    & 39.9      & 36.6        & 38.4 & 33.3    \\
                             & -         & -3.0        & -    & -2.6    & -         & -3.3        & -    & -5.1    \\
\multirow{2}{*}{Structural}  & 37.1      & 35.9        & 45.1 & 43.4    & 50.2      & 47.3        & 51.3 & 48.2    \\
                             & -         & -1.2        & -    & -1.7    & -         & -2.9        & -    & -3.1    \\
\multirow{2}{*}{Explanation} & 10.5      & 9.8         & 7.3  & 3.6     & 16.1      & 12.9        & 15.2 & 10.9    \\
                             & -         & -0.7        & -    & -3.7    & -         & -3.2        & -    & -4.3    \\
\multirow{2}{*}{Knowledge}   & 32.2      & 29.9        & 25.7 & 24.1    & 36.2      & 33.5        & 28.5 & 25.1    \\
                             & -         & -2.3        & -    & -1.6    & -         & -2.7        & -    & -3.4    \\ \hline
\multirow{2}{*}{Average}     & 27.4      & 25.5        & 28.3 & 25.9    & 35.6      & 32.6        & 33.6 & 29.7    \\
                             & -         & -1.9        & -    & -2.4    & -         & -3.0        & -    & -3.9    \\ \hline\hline
\end{tabular}
}
\caption{Evaluation results on InternLM2-chat-7B~(InternLM2 for short), InterLM-XComposer2-4KHD-7B~(InternLM2-X for short), Qwen-chat-7B~(Qwen for short), and Qwen-VL-chat-7B~(Qwen-VL for short) under Q setting and Q+C setting.}
\label{tab:mm}
\end{table*}

\section{More Experiments}~\label{sec:experiment}

In this section, we present additional experimental results that illustrate the performance of MLLMs and their proficiency in leveraging multimodal information within financial contexts. 
In Section~\ref{sec:llmmllm}, we evaluate and compare the performance of Large Language Models (LLMs) and their multimodal counterparts in text-based settings. Subsequently, in Sections~\ref{sec:4compare}, we detail the performance of 13 advanced MLLMs under various input settings, demonstrating their capabilities and limitations in handling complex multimodal data.

\subsection{Large Language Model~(LLMs) \emph{v.s.} Multimodal Large Language Model~(MLLMs)}~\label{sec:llmmllm}

To further understand how Large Language Models (LLMs) and their multimodal counterparts (MLLMs) perform in financial domain, we conduct a study comparing models like Qwen and InternLM2 against their multimodal versions, Qwen-VL-Chat and InternLM-Xcomposer2-4KHD. 
This analysis focused on how these models handle pure text inputs, comparing two settings: setting-Q, which involves only question to evaluate their ability on basic financial knowledge, and setting-Q+C, which includes questions and texture descriptions to evaluate the text processing ability.

The findings, shown in Table~\ref{tab:mm}, reveal that MLLMs tend to perform worse in pure text scenarios compared to their respective LLMs after being trained on multimodal data. 
The performance drop observed in setting-Q suggests that training with multimodal data might lead to some loss of previously acquired financial knowledge in LLMs. 
Moreover, the gap between LLMs and MLLMs widens further in setting-Q+C, indicating that text processing ability of MLLMs has also diminished somewhat compared to LLMs.
This points to a significant challenge for future development of MLLMs, \emph{i.e.}, how to effectively incorporate multimodal data without compromising the quality of text processing. Exploring innovative ways to maintain strong text processing skills while integrating diverse data types, such as through advanced fine-tuning techniques or hybrid architectures, is an important area for further research and development in the field of multimodal learning.

\subsection{More Detailed Results of different evaluation settings}~\label{sec:4compare}

In Table~\ref{tab:results_Q},~\ref{tab:results_Q+C},~\ref{tab:results_Q+I},and~\ref{tab:results_Q+I+C}, we illustrate the performance of 13 MLLMs across four experimental settings: question only, question with image, question with caption, and question with both image and caption, demonstrating their efficacy in multimodal financial contexts. The tables record the evaluation results of 13 models under different settings, sorted in ascending order based on the average values.
All open-source MLLMs achieve their best results when utilizing a combination of questions, images, and captions as inputs. 
This is primarily because the captions, generated by GPT-4V, often provide additional context that is not readily apparent from the images alone.
However, as these captions are generated by GPT-4V and only encapsulate a subset of the information presented in the charts, they do not enhance the performance of GPT-4V. Nevertheless, GPT-4V still secures the top position among all models with its exceptional comprehension capabilities in the first three settings, while in the fourth setting, CogVLM2 demonstrates even more remarkable prowess.
Furthermore, when comparing the outcomes between setting-Q+I (question and image as input) and setting-Q+C (question and caption as input), the results generally indicate a slight superiority of setting-Q+I over setting-Q+C. This comparison underscores the robust capability of MLLMs to leverage multimodal data effectively.

Experimental results indicate that while these cutting-edge MLLMs are progressing impressively, there remains substantial potential for enhancement in their performance within complex multimodal financial applications. This observation underscores the ongoing need for further development and refinement in the advanced capabilities of MLLMs to effectively integrate, analyze, and utilize diverse multimodal data in the financial domain.

\begin{table*}[htbp]
\centering
\caption{The specific results of 13 cutting-edge MLLMs on CFBenchmark-MM with question as only input (setting-Q).}
\label{tab:results_Q}
\resizebox{\textwidth}{!}{
\begin{tabular}{c|ccccc|c}
\toprule
\textbf{Models} & \textbf{Statistical} & \textbf{Arithmetic} & \textbf{Structural} & \textbf{Explanation} & \textbf{Knowledge} & \textbf{Average}\\
\hline
LLaVA-NeXT-mistral-7B & 16.6 & 18.2 & 17.4 & 2.2 & 15.4 & 14.0 \\
LLaVA-NeXT-vicuna-13B & 22.6 & 19.1 & 27.2 & 4.0 & 21.3 & 18.8 \\
LLaVA-NeXT-vicuna-7B & 29.1 & 28.4 & 30.5 & 6.3 & 10.4 & 21.0 \\
mPLUG-0w1 & 24.8 & 22.2 & 29.0 & 7.0 & 25.9 & 21.8 \\
Llama-3.2-11B & 25.3 & 29.3 & 27.7 & 8.1 & 28.4 & 23.8 \\
Baichuan-Omni-1d5 & 27.9 & 25.7 & 28.8 & 12.6 & 25.9 & 24.2 \\
CogVLM2 & 30.1 & 29.2 & 34.7 & 7.6 & 24.5 & 25.2 \\
Qwen2.5-VL-7B-Instruct & 29.3 & 27.5 & 30.2 & 10.2 & 29.4 & 25.3 \\
InternLM-XComposer2-4KHD-7B & 27.4 & 24.4 & 35.9 & 9.8 & 29.9 & 25.5 \\
DeepSeek-VL-7B & 30.9 & 29.1 & 39.2 & 5.4 & 30.9 & 27.1 \\
InternLM-XComposer2-VL-7B & 30.4 & 29.1 & 39.5 & \textbf{12.7} & \textbf{33.2} & 29.0 \\
LLaVA-NeXT-34B & 36.2 & \textbf{34.0} & 40.8 & 6.3 & 29.0 & 29.3 \\
GPT-4V & \textbf{38.0} & 26.9 & \textbf{41.9} & 8.2 & 31.8 & \textbf{29.4} \\
\bottomrule
\end{tabular}
}
\end{table*}

\begin{table*}[htbp]
\centering
\caption{The specific results of 13 cutting-edge MLLMs on CFBenchmark-MM with  question and caption as input (setting-Q+C).}
\label{tab:results_Q+C}
\resizebox{\textwidth}{!}{
\begin{tabular}{c|ccccc|c}
\toprule
\textbf{Models} & \textbf{Statistical} & \textbf{Arithmetic} & \textbf{Structural} & \textbf{Explanation} & \textbf{Knowledge} & \textbf{Average} \\
\hline
LLaVA-NeXT-mistral-7B & 20.8 & 22.4 & 25.4 & 7.7 & 18.3 & 18.9 \\
LLaVA-NeXT-vicuna-7B & 33.5 & 29.9 & 35.0 & 14.9 & 13.9 & 25.4 \\
mPLUG-0w1 & 29.9 & 26.7 & 33.8 & 11.1 & 28.4 & 26.0 \\
LLaVA-NeXT-vicuna-13B & 33.1 & 28.3 & 38.6 & 12.5 & 25.6 & 27.6 \\
Baichuan-Omni-1d5 & 36.2 & 33.1 & 39.8 & 13.1 & 35.9 & 31.6 \\
CogVLM2 & 35.2 & 39.4 & 37.9 & 18.3 & 28.2 & 31.8 \\
Llama-3.2-11B & 40.4 & 36.1 & 39.2 & 10.7 & 33.5 & 32.0 \\
InternLM-XComposer2-4KHD-7B & 36.6 & 32.8 & 47.3 & 13.0 & 33.5 & 32.6 \\
Qwen2.5-VL-7B-Instruct & 38.8 & 34.2 & 40.9 & 15.5 & 36.3 & 33.1 \\
InternLM-XComposer2-VL-7B & 35.1 & 33.1 & 44.9 & 13.8 & 38.8 & 33.2 \\
LLaVA-NeXT-34B & 40.3 & \textbf{38.0} & 47.1 & 13.6 & 34.6 & 34.7 \\
DeepSeek-VL-7B & 40.4 & 34.0 & 46.1 & \textbf{15.2} & \textbf{39.6} & 35.1 \\
GPT-4V & \textbf{41.5} & 35.4 & \textbf{50.0} & 11.6 & 38.2 & \textbf{35.3} \\
\bottomrule
\end{tabular}
}
\end{table*}

\begin{table*}[htbp]
\centering
\caption{The specific results of 13 cutting-edge MLLMs on CFBenchmark-MM with  question and image as input (setting-Q+I).}
\label{tab:results_Q+I}
\resizebox{\textwidth}{!}{
\begin{tabular}{c|ccccc|c}
\toprule
\textbf{Models} & \textbf{Statistical} & \textbf{Arithmetic} & \textbf{Structural} & \textbf{Explanation} & \textbf{Knowledge} & \textbf{Average} \\
\hline
LLaVA-NeXT-mistral-7B & 22.4 & 24.4 & 26.6 & 5.6 & 18.1 & 19.4 \\
LLaVA-NeXT-vicuna-7B & 31.8 & 32.3 & 30.8 & 11.3 & 15.2 & 24.3 \\
LLaVA-NeXT-vicuna-13B & 29.6 & 26.7 & 29.9 & 12.3 & 26.3 & 25.0 \\
mPLUG-0w1 & 28.0 & 25.2 & 31.4 & 12.1 & 30.9 & 25.5 \\
DeepSeek-VL-7B & 35.4 & 31.8 & 44.9 & 12.6 & 39.3 & 32.8 \\
InternLM-XComposer2-VL-7B & 38.6 & 33.0 & 46.7 & 21.7 & 41.6 & 36.3 \\
LLaVA-NeXT-34B & 42.8 & 39.2 & 50.3 & 15.8 & 44.2 & 38.4 \\
Baichuan-Omni-1d5 & 42.5 & 38.6 & 46.7 & 21.0 & 46.6 & 39.1 \\
InternLM-XComposer2-4KHD-7B & 44.6 & 38.9 & 52.1 & 27.7 & 35.5 & 39.8 \\
Qwen2.5-VL-7B-Instruct & 43.1 & 40.7 & 49.2 & 22.1 & 49.5 & 40.9 \\
Llama-3.2-11B & 45.1 & 42.3 & 52.9 & 18.1 & 46.5 & 41.0 \\
CogVLM2 & 46.6 & 40.6 & 55.4 & \textbf{34.2} & 38.6 & 43.1 \\
GPT-4V & \textbf{52.0} & \textbf{46.0} & \textbf{58.1} & 24.8 & \textbf{52.4} & \textbf{46.7} \\
\bottomrule
\end{tabular}
}
\end{table*}

\begin{table*}[htbp]
\centering
\caption{The specific results of 13 cutting-edge MLLMs on CFBenchmark-MM with  question , image and caption as input (setting-Q+I+C).}
\label{tab:results_Q+I+C}
\resizebox{\textwidth}{!}{
\begin{tabular}{c|ccccc|c}
\toprule
\textbf{Models} & \textbf{Statistical} & \textbf{Arithmetic} & \textbf{Structural} & \textbf{Explanation} & \textbf{Knowledge} & \textbf{Average} \\
\hline
LLaVA-NeXT-mistral-7B & 26.2 & 25.0 & 33.2 & 11.2 & 11.3 & 21.4 \\
mPLUG-0w1 & 30.5 & 26.4 & 33.5 & 14.4 & 31.1 & 27.2 \\
LLaVA-NeXT-vicuna-7B & 36.4 & 33.7 & 39.8 & 13.6 & 18.1 & 28.3 \\
LLaVA-NeXT-vicuna-13B & 33.7 & 29.0 & 40.7 & 15.8 & 29.7 & 29.8 \\
DeepSeek-VL-7B & 39.1 & 34.0 & 46.7 & 13.4 & 41.1 & 34.9 \\
InternLM-XComposer2-VL-7B & 41.4 & 35.7 & 46.1 & 21.4 & 42.4 & 37.4 \\
LLaVA-NeXT-34B & 45.2 & 42.1 & 55.7 & 17.1 & 48.3 & 41.7 \\
InternLM-XComposer2-4KHD-7B & 45.9 & 40.2 & 54.2 & 31.2 & 38.9 & 42.1 \\
GPT-4V & 45.2 & 38.1 & 54.6 & 23.8 & 49.9 & 42.3 \\
Baichuan-Omni-1d5 & 43.3 & 42.7 & 51.9 & 24.8 & \textbf{50.1} & 42.6 \\
Qwen2.5-VL-7B-Instruct & 46.3 & 45.1 & 53.2 & 26.1 & 51.6 & 44.5 \\
Llama-3.2-11B & \textbf{49.3} & \textbf{47.1} & 54.9 & 22.9 & 49.3 & 44.7 \\
CogVLM2 & 48.5 & 41.9 & \textbf{58.0} & \textbf{35.9} & 40.1 & \textbf{44.9} \\

\bottomrule
\end{tabular}
}
\end{table*}

\section{More Examples}~\label{sec:example}

\subsection{Benchmark Examples}

\begin{figure*}[ht]
    \centering
    \includegraphics[width=1\linewidth]{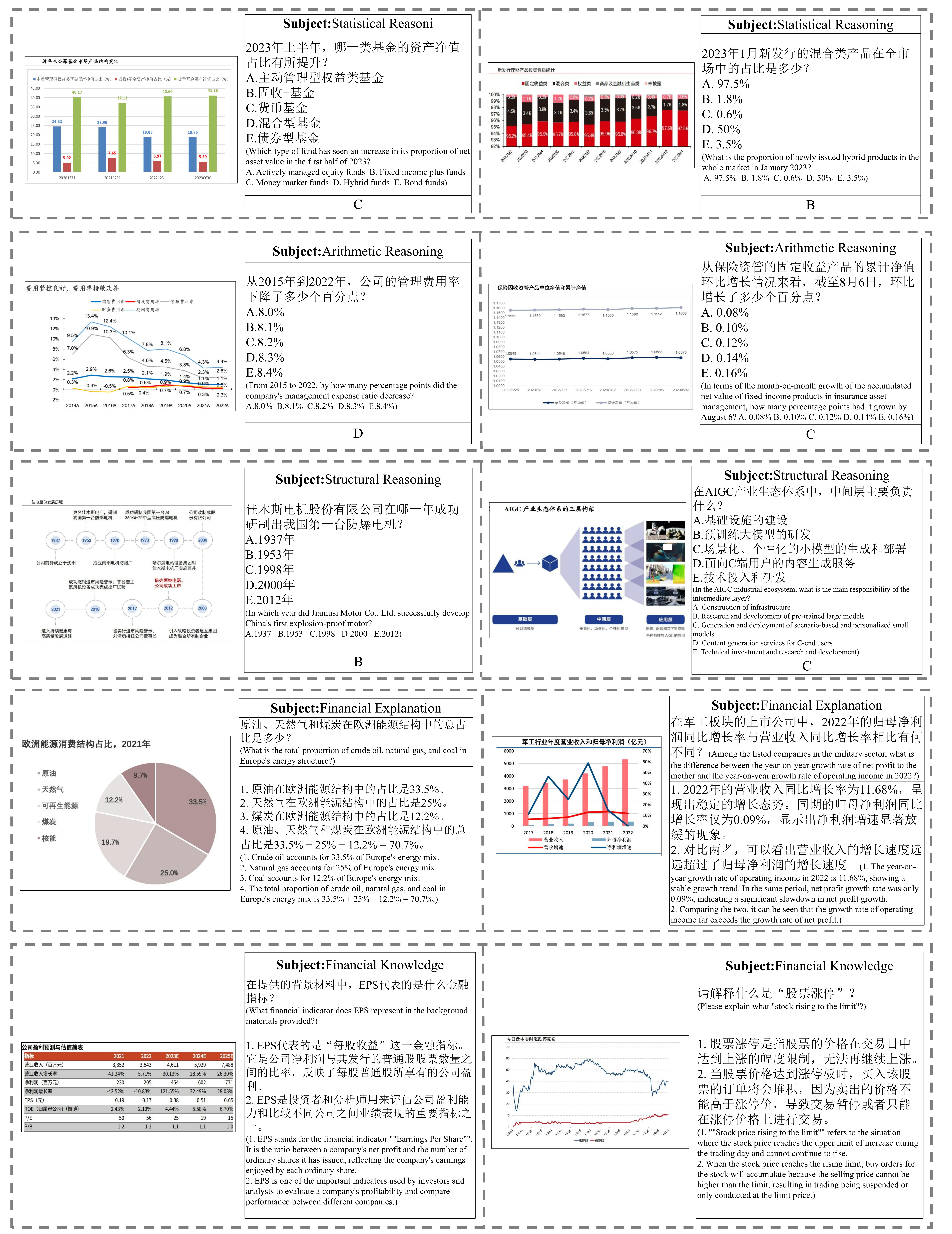}
    \caption{Ten examples from CFBenchmark-MM}
    \label{fig:cases2}
\end{figure*}

In Figure~\ref{fig:cases2}, we showcase ten examples from the CFBenchmark-MM, covering a spectrum of tasks including arithmetic reasoning, statistical reasoning, structural reasoning, financial explanation, and financial knowledge. 
It encompasses a relatively comprehensive range of chart types, including bar charts, line graphs, structural diagrams, pie charts, tables, and more, demonstrating the various ways in which financial question-and-answer constructs can be built across different categories. 
Following the publication of this paper, the benchmark will be released, comprising only the questions, images, and captions to preclude the possibility of training the MLLMs directly on CFBenchmark-MM.

\subsection{Failure Cases}

\begin{figure*}
    \centering
    \includegraphics[width=1.\linewidth]{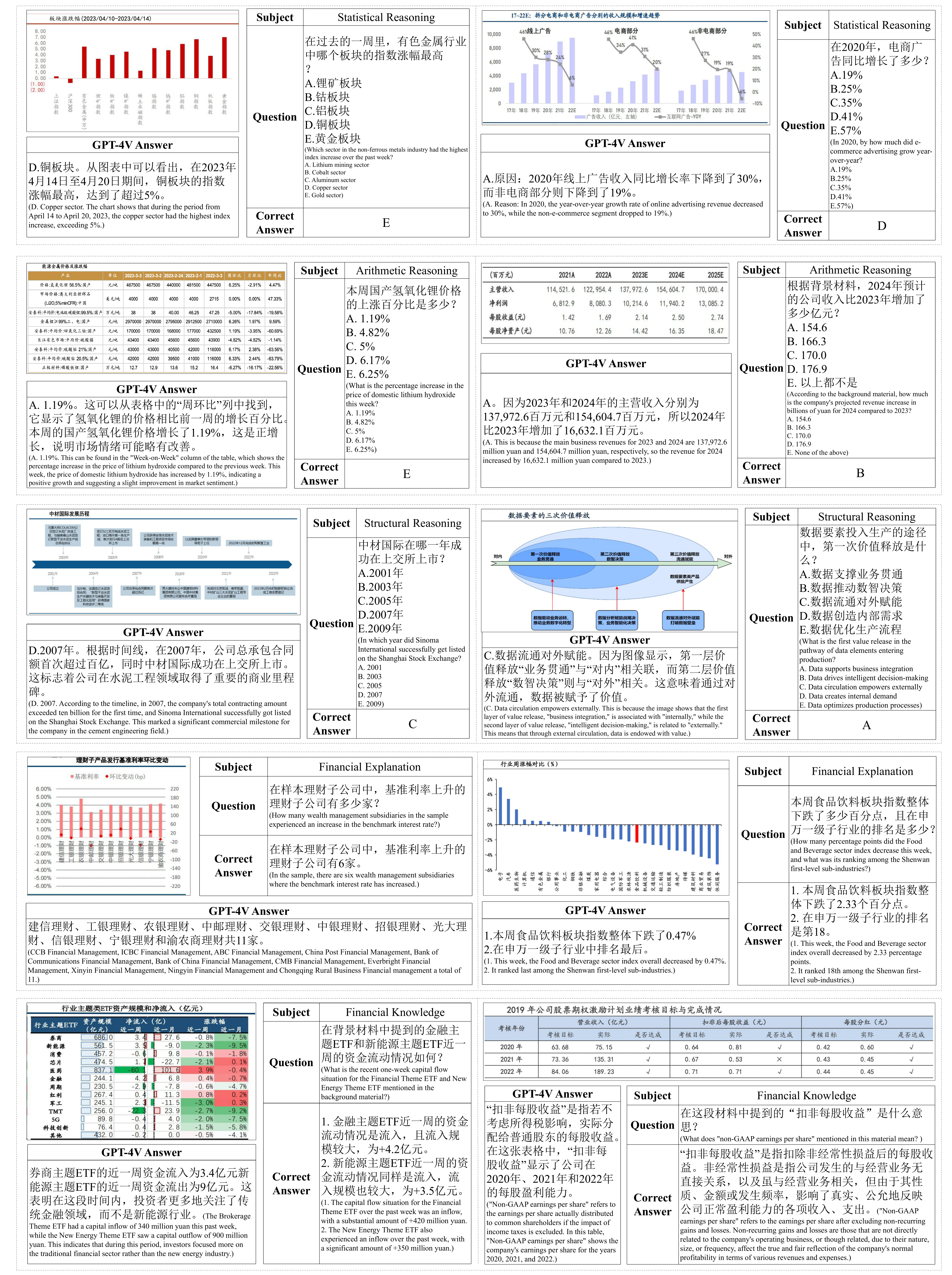}
    \caption{Failure cases in CFBenchmark-MM by GPT-4V with reasons and answers.}
    \label{fig:fail_case}
\end{figure*}

In Figure~\ref{fig:fail_case}, we present additional examples to illustrate and analyze the challenges faced by MLLMs in multimodal financial domain, focusing on failures of GPT-4V during the answer process. 
As mentioned in our main submission, the primary issues involve the misinterpretation of visual content and misunderstanding of financial concepts.

One notable failure, showcased in the first row, second column of Figure~\ref{fig:fail_case}, reveals that the model struggles to effectively correlate the text mentioning ``e-commerce advertising'' with the visual segment labeled as ``e-commerce sector'' in the chart. This misalignment prevents the model from deriving the correct answer.
Another example, displayed in the second row, first column, highlights the limited ability of GPT-4V to interpret visual data accurately. Although it recognizes the need to locate lithium hydroxide in the week-on-week column, its inadequate image processing capabilities lead to an inability to pinpoint the correct figure of 6.25\%.
Addressing these issues by enhancing the interpretation of visual content and deepening the understanding of financial concepts is crucial for improving the application of MLLMs in financial contexts. Effective strategies to overcome these challenges will significantly advance the utility of MLLMs in complex financial environments.

\subsection{Self Correction Cases}

In the evaluation process of MLLMs, GPT-4V has demonstrated significant self-correction capabilities. 
Specifically, when GPT-4V initially provides an incorrect response, it is able to revise its answer appropriately if the reasons for the errors are clearly indicated. 
This capacity for adjustment enhances its utility in dynamic application.

\begin{figure*}
    \centering
    \includegraphics[width=0.95\linewidth]{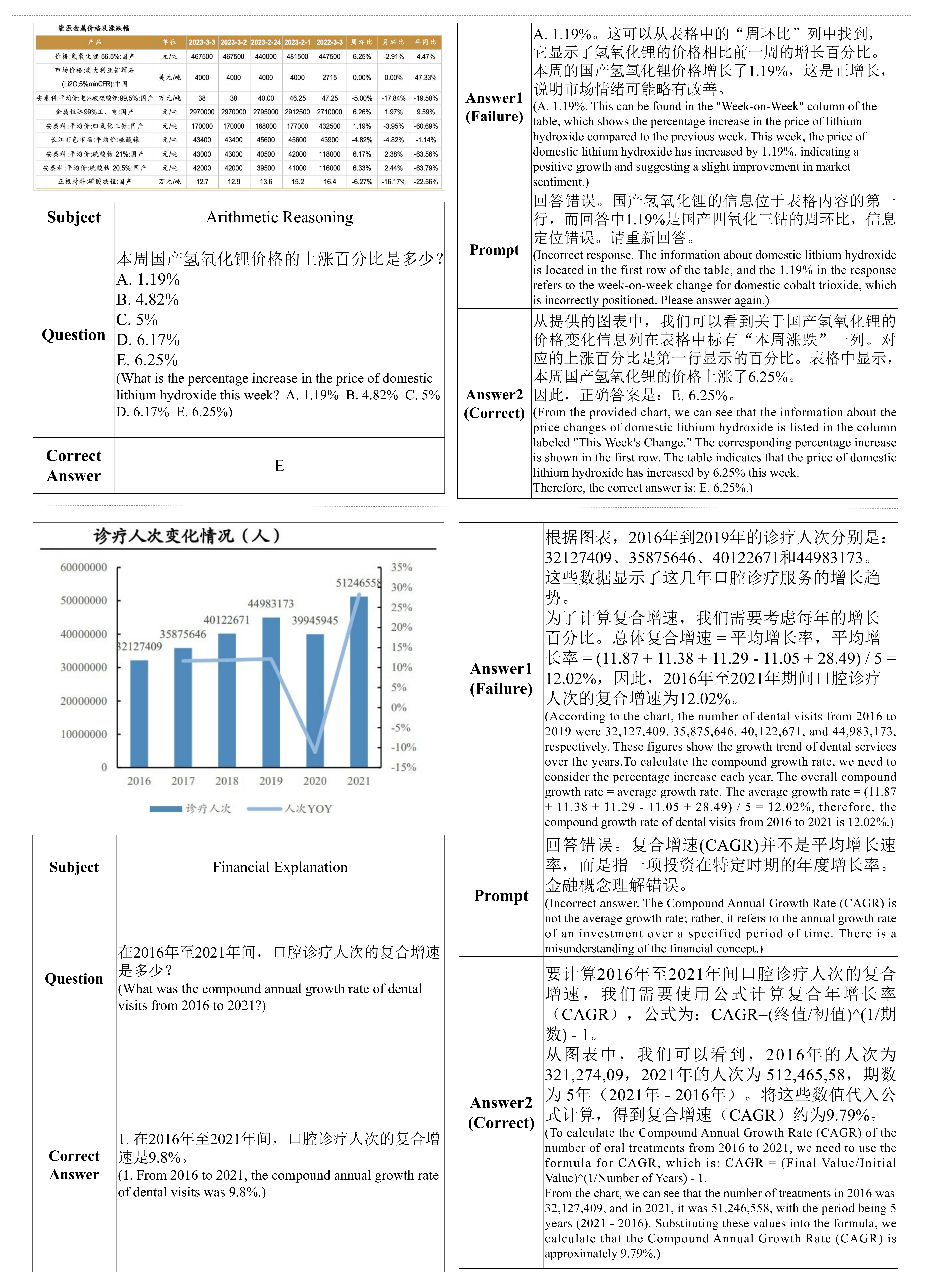}
    \caption{Self-correction cases in CFBenchmark-MM generated by GPT-4V.}
    \label{fig:self_correction}
\end{figure*}

Figure~\ref{fig:self_correction} illustrates two instances of this self-correction process. 
In the first example, we highlight that GPT-4V does not accurately locate the week-on-week percentage change for lithium hydroxide. 
After the error is pointed out, GPT-4V successfully identify and correct its mistake in a subsequent interaction. 
In the second case, the model initially confuses the terms `compound growth rate' and `average growth rate.' Once these errors are clarified, GPT-4V effectively adjusts its response to provide the correct information in the following round of dialogue.
Such robust self-correction capabilities, as observed in GPT-4V, have not been evident in tests involving other open-source models. 
This unique ability of GPT-4V not only underscores its advanced functionality but also provides valuable insights for the future design of more accurate multi-agent reasoning systems. 
This capability is instrumental in advancing MLLMs in complex financial application, where error correction plays a critical role in maintaining the reliability and accuracy of model outputs.

\end{document}